\documentclass{article}

\PassOptionsToPackage{numbers, compress}{natbib}

\usepackage[preprint]{neurips_2026}

\usepackage{hyperref}       

\hypersetup{
    colorlinks=true,
    linkcolor=red,
    citecolor=blue,
    urlcolor=blue,
}

\setcitestyle{numbers,square,comma}

\usepackage{url}            
\usepackage{booktabs}       
\usepackage{amsfonts}       
\usepackage{nicefrac}       
\usepackage{microtype}      
\usepackage{xcolor}         
\usepackage{multirow}       
\usepackage{algorithm}      
\usepackage{algorithmic}    
\usepackage{amsmath}        
\usepackage{graphicx}
\usepackage{tcolorbox}
\usepackage{tabularx} 
\usepackage{subcaption}
\usepackage{caption}
\usetikzlibrary{positioning,fit,shapes.geometric}

\title{FlashPDE: A Drop-In Fused Triton Operator Library for Neural PDE Solvers}

\author{
Peiyu Zang$^{1}$,
Bosen Xie$^{2}$,
Ruoxiang Xu$^{1}$,
Yongqiang Cai$^{1,*}$\\
$^{1}$School of Mathematical Sciences, Beijing Normal University\\
$^{2}$Beijing University of Posts and Telecommunications\\
\texttt{202421130105@mail.bnu.edu.cn}\\
\texttt{caiyq.math@bnu.edu.cn}
}

\begin{document}

\maketitle

\begin{abstract}
Training Physics-Informed Neural Networks (PINNs) typically relies on automatic differentiation to compute high-order PDE residuals, a process that retains nested computation graphs and causes substantial memory overhead on 3D domains. While output-grid finite-difference (FD) approximations eliminate these tracking graphs and reduce memory footprints, native PyTorch implementations suffer from kernel launch fragmentation, incurring substantial memory-bound I/O and dispatch overhead. We present \textbf{FlashPDE}, a fused-kernel operator library for grid-based SciML. FlashPDE provides 14 differentiable PDE operators implemented with hand-derived discrete adjoints and custom Triton kernels, covering 17 benchmark configurations across diverse 1D--3D scalar and multi-field PDEs. Each operator is exposed as an architecture-independent, drop-in \texttt{torch.autograd.Function} that consumes grid-aligned physical fields from compatible PyTorch models and uses three Triton launches: a fused forward stencil, an analytic-adjoint backward kernel, and a boundary-gradient correction kernel. The same operators support MLP and CNN field generators without changes to the operator implementation, outer loss construction, or boundary conditions. We evaluate six representative main cases in detail and report supplementary results across all 17 configurations. FlashPDE achieves median T2S speedups of up to \textbf{2.30$\times$} over the eager PyTorch FD baseline, while maintaining comparable case-specific accuracy metrics across backends, and up to \textbf{19.2$\times$} kernel-level acceleration on the largest evaluated grids. The FD discretization reduces memory of complex four-field fluid simulations (\texttt{tgv\_3d}) from $>40$\,GB under vanilla autograd to 3.4\,GB, which FlashPDE maintains while providing the throughput speedup. FlashPDE extends Triton kernel fusion to differentiable PDE stencil workloads, connecting fused GPU execution with the native PyTorch ecosystem.
\end{abstract}

\section{Introduction}
\label{sec:intro}

Physics-Informed Neural Networks (PINNs) and broader scientific machine
learning (SciML) methodologies solve partial differential equations (PDEs) by
incorporating physical constraints into neural-network training through
residual minimization~\cite{raissi2019physics,lu2021deepxde,
krishnapriyan2021characterizing,cuomo2022scientific}. 
With the development of neural operators and learned physical
models~\cite{lu2021deeponet,li2021fourier,raissi2020hidden,
sanchez2020learning}, PyTorch~\cite{paszke2019pytorch} has
become a widely adopted platform for SciML research. However, integrating
deep-learning frameworks with high-performance numerical computation remains a
fundamental challenge. Traditional HPC stencil systems achieve high throughput
through explicit memory-access control and hardware-aware
scheduling~\cite{sai2022accelerating,louboutin2019devito,gysi2015stela}.
Combining such optimized numerical kernels with differentiable neural training
requires bridging the gap between framework-level flexibility and hardware-level
efficiency.

Inside PyTorch, evaluating higher-order PDE derivatives introduces two
fundamental bottlenecks:
 \begin{itemize}
    \item \textbf{Autograd Memory Overhead:}
    Standard PINN implementations rely on
    \texttt{torch.autograd} to compute spatial derivatives. Higher-order
    derivatives require retaining nested computation graphs through
    \texttt{create\_graph=True}, resulting in substantial memory overhead from
    derivative-related intermediate tensors~\cite{lu2021deepxde}. For large three-dimensional
    PDE systems, this memory overhead can become the dominant limitation and
    prevent training on available GPU resources.

    \item \textbf{Kernel Launch Fragmentation:}
    CAN-PINN~\cite{canpinn} reduces reliance on purely coordinate-based
    automatic differentiation by coupling automatic and numerical
    differentiation. Output-grid finite-difference formulations provide a
    separate alternative: they evaluate PDE derivatives directly from
    neighboring network outputs, avoiding nested coordinate-derivative graphs
    and substantially reducing memory usage. However,
    native PyTorch FD implementations express stencils through tensor slicing
    and element-wise operations, causing many independent CUDA kernel launches
    and repeated global-memory traffic. Existing compilation systems improve
    tensor execution in many deep-learning workloads~\cite{ansel2024torchcompile}, but structured-grid PDE
    operators with multi-field coupling and complex stencil dependencies are
    not automatically fused in typical eager execution workflows.
\end{itemize}

For example, in the evaluated 3D TGV case, coordinate autograd exceeds the
40\,GB device capacity, while output-grid FD reduces peak allocation to
3.39\,GB. However, a profiled forward--loss--backward iteration using eager
PyTorch FD still records 934 CUDA kernel launches, compared with 267 for
FlashPDE.

\begin{figure*}[t]
    \centering
    \includegraphics[width=\textwidth]{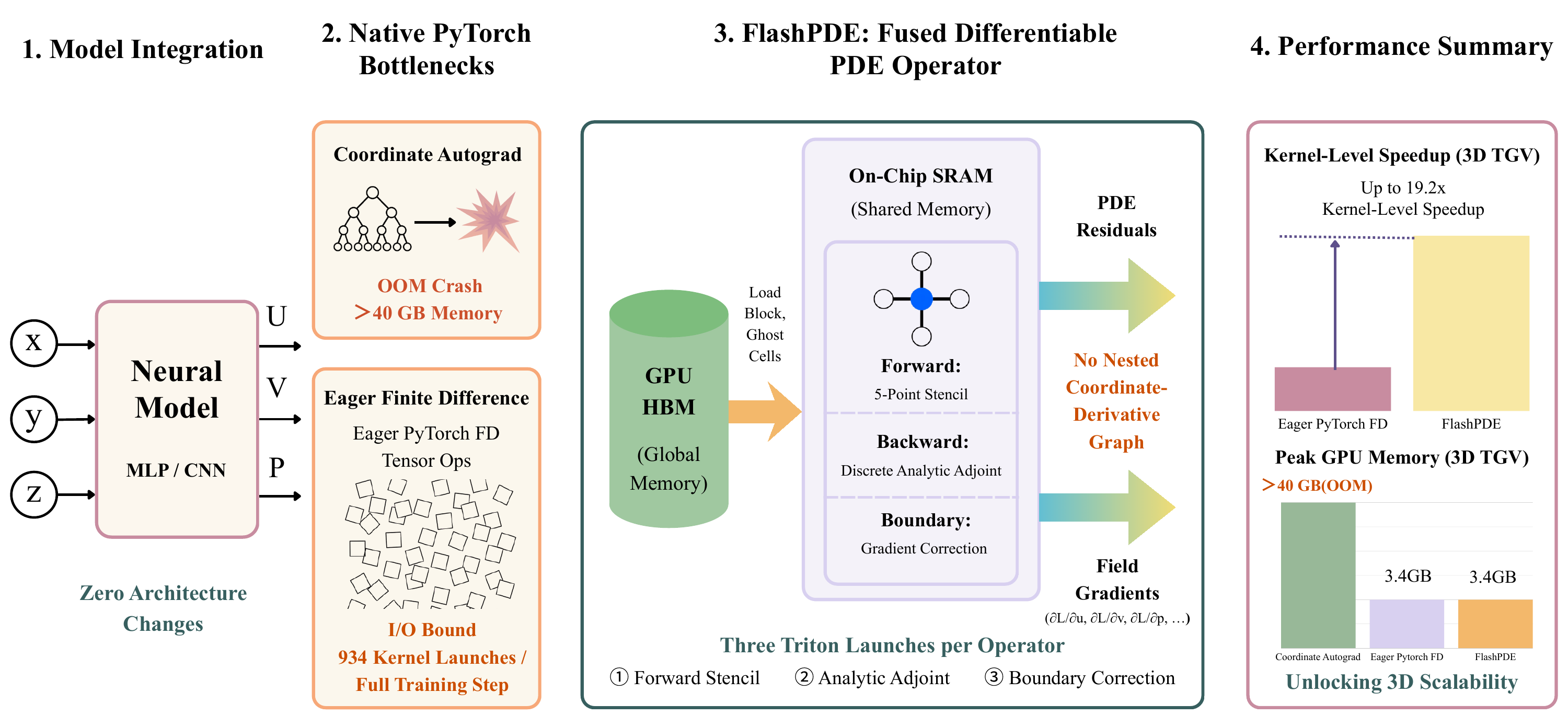}
\caption{
Overview of FlashPDE.
(a) Drop-in integration.
(b) Native PyTorch bottlenecks.
(c) Fused differentiable operators.
(d) Performance improvements.
}
    \label{fig:overview}
\end{figure*}

To address this gap, we propose \textbf{FlashPDE} (see Figure~\ref{fig:overview}), a fused-kernel operator
library for grid-based SciML. FlashPDE replaces PyTorch-native PDE residual
evaluation with custom differentiable operators implemented through Triton GPU
kernels~\cite{tillet2019triton}. Each operator combines the forward finite-
difference stencil, an analytic discrete-adjoint backward pass, and boundary
gradient corrections inside a unified
\texttt{torch.autograd.Function}. The current library contains 14 differentiable PDE operators implemented with
hand-derived discrete adjoints and custom Triton kernels, supporting 17 PDE
configurations across 1D--3D elliptic, parabolic, hyperbolic, conservation-law,
phase-field, and incompressible/compressible-flow problems.

FlashPDE operates as an architecture-independent execution layer between
neural field generators and PDE residual evaluation. Because its operators
consume grid-aligned physical field tensors rather than network internals, the
same implementation can be used with compatible MLP, CNN, or other PyTorch
field generators without modifying the neural architecture, optimization
procedure, or boundary-condition formulation.

Developing FlashPDE requires addressing the challenge of differentiating
through fused nonlinear PDE operators. While forward stencil evaluation only
requires local neighborhood accesses, backward propagation involves the
transpose discrete Jacobian of the residual operator. This becomes particularly
challenging for nonlinear and multi-field equations, where one variable can
affect multiple neighboring residual terms. FlashPDE resolves this issue by
deriving analytic adjoint formulations and implementing the complete
forward--backward computation within fused Triton kernels.

Our contributions are summarized as follows:

\begin{enumerate}
    \item \textbf{Architecture-Independent Differentiable PDE Operator Library:}
    We develop FlashPDE, a collection of 14 drop-in differentiable PDE operators
    covering 17 SciML PDE configurations. Each operator consumes grid-aligned
    physical field tensors independently of the neural field generator and
    integrates forward FD evaluation, discrete adjoint backward propagation,
    and boundary-gradient correction within a unified PyTorch interface. The
    same operator implementations can therefore be reused across compatible
    MLP and CNN models.

    \item \textbf{Memory and Execution Attribution:}
    We provide a component-level analysis separating the memory benefits of
    finite-difference formulations from the execution improvements introduced
    by GPU kernel fusion. FlashPDE preserves the low-memory characteristics of
    FD while reducing the kernel-launch and intermediate-memory overhead of
    eager PyTorch implementations.

    \item \textbf{Performance and Accuracy Evaluation:}
    We evaluate FlashPDE across six representative PDE benchmarks spanning
    1D--3D problems and Navier--Stokes systems. FlashPDE achieves up to
    $2.30\times$ end-to-end time-to-solution speedup over eager PyTorch FD
    baselines, with kernel-level acceleration reaching $19.2\times$ on
    the largest evaluated grids.
\end{enumerate}

\section{FlashPDE Kernel Design and Discrete Analytic Adjoints}
\label{sec:kernel}

FlashPDE replaces the eager PyTorch evaluation of supported
finite-difference (FD) residual operators with custom Triton operators exposed
through \texttt{torch.autograd.Function}. Each operator integrates three
components: a fused forward stencil evaluation, a discrete analytic-adjoint
backward pass, and boundary-gradient corrections. The operator consumes physical
field tensors defined on regular Cartesian grids and is independent of the neural
architecture used to generate these fields.

\subsection{Fused Forward Stencil}
\label{sec:forward_kernel}

Consider the steady two-dimensional incompressible Navier--Stokes equations. At
an interior grid point $(i,j)$, the momentum and continuity residuals are
\begin{align*}
    \mathrm{res}_u^{i,j}
    &=
    u^{i,j}u_x^{i,j}
    +
    v^{i,j}u_y^{i,j}
    +
    p_x^{i,j}
    -
    \nu\left(u_{xx}^{i,j}+u_{yy}^{i,j}\right),
    \\
    \mathrm{res}_v^{i,j}
    &=
    u^{i,j}v_x^{i,j}
    +
    v^{i,j}v_y^{i,j}
    +
    p_y^{i,j}
    -
    \nu\left(v_{xx}^{i,j}+v_{yy}^{i,j}\right),
    \\
    \mathrm{res}_{\mathrm{div}}^{i,j}
    &=
    u_x^{i,j}+v_y^{i,j}.
\end{align*}

An eager PyTorch FD implementation evaluates these derivatives and residual
expressions through tensor slicing and element-wise operations. Although tensor
views may avoid explicit copies, subsequent arithmetic operations are executed
as independent CUDA kernels and may materialize intermediate derivative tensors
in global memory. As the PDE residual becomes more complex, this execution
pattern introduces substantial kernel-launch and memory-access overhead
(Appendix~\ref{app:code_comparison}).

FlashPDE assigns each Triton program instance to a spatial tile. Required field
values are loaded through coalesced memory accesses, while centered derivatives,
nonlinear terms, and residual expressions are evaluated within the same kernel.
Only the final residual fields are written back to global memory. Therefore, the
complete forward PDE residual evaluation is reduced to a single Triton launch
with autotuned block configuration (Appendix~\ref{app:autotune}).
Representative forward pseudocode is provided in
Appendix~\ref{app:algorithms}.

Figure~\ref{fig:fd_vs_flashpde} contrasts the two execution paths.
Table~\ref{tab:launch_memory} summarizes the execution and memory
characteristics of the six main cases. FlashPDE reduces the number of CUDA
kernel launches by $1.7\times$--$3.5\times$ relative to eager PyTorch FD.

\begin{figure}[t]
\centering
\includegraphics[width=\linewidth]{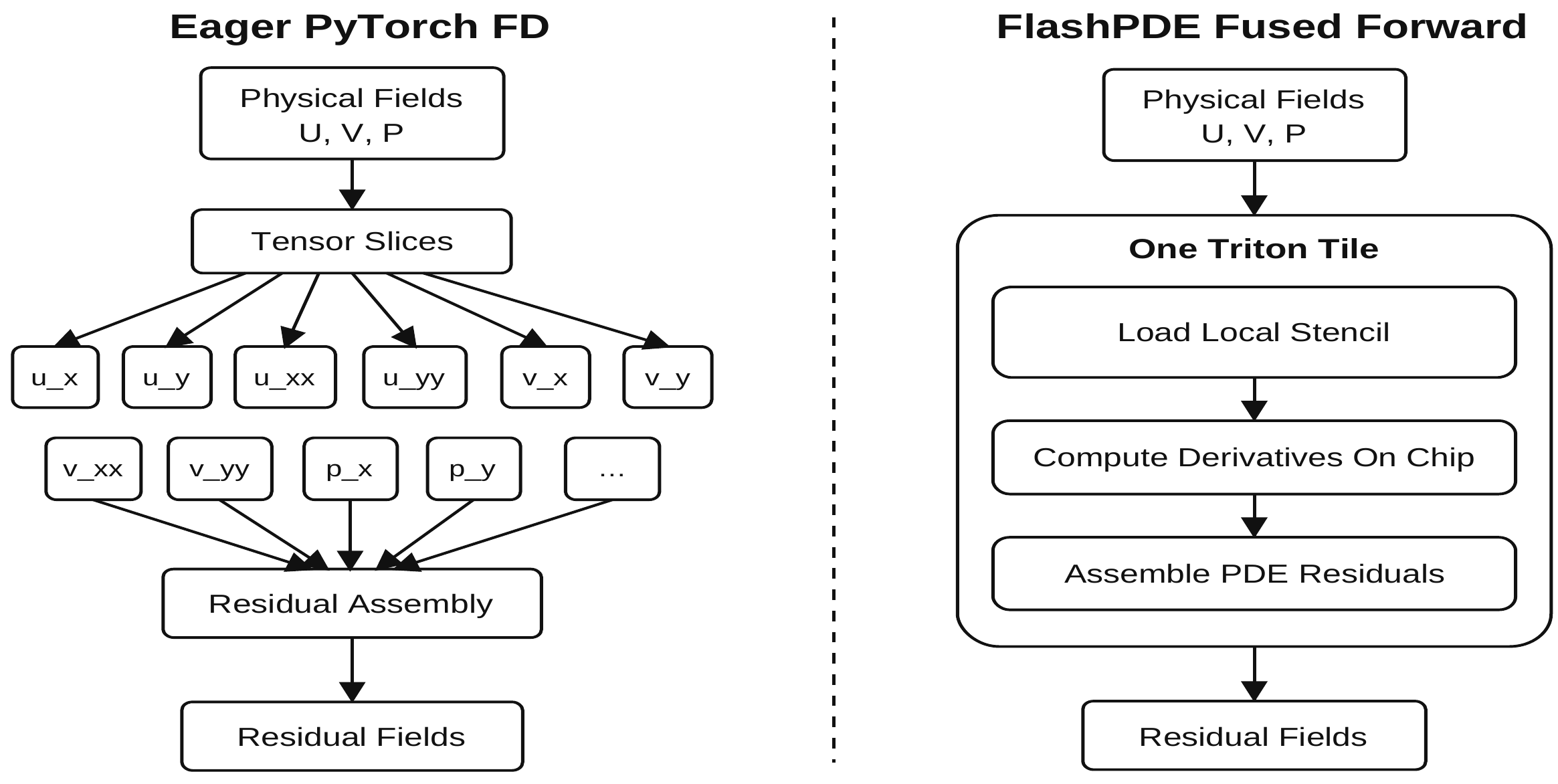}
\caption{
Execution comparison between eager PyTorch finite-difference operators and
FlashPDE fused Triton kernels. FlashPDE fuses stencil evaluation and PDE
residual computation into one forward kernel.
}
\label{fig:fd_vs_flashpde}
\end{figure}

\begin{table*}[t]
\centering
\caption{
Execution and memory comparison across the six main cases. CUDA kernel-launch
reduction compares FlashPDE with eager PyTorch FD, whereas memory reduction
compares output-grid FD with coordinate-based automatic differentiation.
Launch counts are measured per forward--loss--backward iteration, excluding the
optimizer update; OOM indicates that coordinate autograd exceeds the 40\,GB
device capacity.
}
\label{tab:launch_memory}
\small
\setlength{\tabcolsep}{6pt}
\begin{tabular}{lrrcrrc}
\toprule
& \multicolumn{3}{c}{\textbf{CUDA Kernel Launches}}
& \multicolumn{3}{c}{\textbf{Peak GPU Memory}} \\
\cmidrule(lr){2-4}
\cmidrule(lr){5-7}
\textbf{Case}
& \textbf{PyTorch FD}
& \textbf{FlashPDE}
& \textbf{Reduction}
& \textbf{Autograd (GB)}
& \textbf{FD (GB)}
& \textbf{Reduction} \\
\midrule
1D Burgers   & 129 & 76  & $1.7\times$ & 1.075 & 0.183 & $5.9\times$ \\
2D Diffusion & 192 & 84  & $2.3\times$ & 3.254 & 0.356 & $9.1\times$ \\
2D LDC       & 321 & 137 & $2.3\times$ & 1.321 & 0.086 & $15.4\times$ \\
2D TGV       & 452 & 182 & $2.5\times$ & 7.830 & 0.399 & $19.6\times$ \\
3D LDC       & 727 & 231 & $3.1\times$ & 17.59 & 0.47  & $37.0\times$ \\
3D TGV       & 934 & 267 & $3.5\times$ & OOM   & 3.39  & --- \\
\bottomrule
\end{tabular}
\vspace{2pt}

{\footnotesize
Additional results for all 17 PDE configurations are reported in
Appendices~\ref{app:profiler} and~\ref{app:memory}.}
\end{table*}

\subsection{Discrete Analytic-Adjoint Backward}
\label{sec:backward_kernel}

The backward computation is the main implementation challenge in constructing
fused differentiable PDE operators. During \texttt{loss.backward()}, the custom
operator receives the upstream residual sensitivities

\[
G_u
=
\frac{\partial\mathcal{L}}{\partial\mathrm{res}_u},
\qquad
G_v
=
\frac{\partial\mathcal{L}}{\partial\mathrm{res}_v},
\qquad
G_{\mathrm{div}}
=
\frac{\partial\mathcal{L}}
{\partial\mathrm{res}_{\mathrm{div}}}.
\]

Discrete-adjoint methods compute the action of the transpose discrete
Jacobian~\cite{farrell2013dolfinadjoint}. FlashPDE computes this action for the
PDE residual operator with respect to its input physical fields.

For linear self-adjoint operators, the backward operation may share similar
structures with the forward stencil. However, nonlinear multi-field systems
require explicitly accumulating all local, neighboring, and cross-variable
dependencies. In the two-dimensional Navier--Stokes example, the value
$u^{i,j}$ contributes not only to the local horizontal momentum residual, but
also to neighboring momentum equations and continuity constraints. Therefore,
the backward operator cannot be obtained by simply reusing the forward stencil.

Applying the chain rule to the centered finite-difference residual gives the
following discrete adjoint expression for the gradient with respect to
$u^{i,j}$:

\begin{align}
\frac{\partial\mathcal{L}}{\partial u^{i,j}}
={}&
G_u^{i,j}
\left(
    u_x^{i,j}
    +
    \frac{2\nu}{\Delta x^2}
    +
    \frac{2\nu}{\Delta y^2}
\right)
\nonumber\\
&+
G_u^{i-1,j}
\left(
    \frac{u^{i-1,j}}{2\Delta x}
    -
    \frac{\nu}{\Delta x^2}
\right)
+
G_u^{i+1,j}
\left(
    -\frac{u^{i+1,j}}{2\Delta x}
    -
    \frac{\nu}{\Delta x^2}
\right)
\nonumber\\
&+
G_u^{i,j-1}
\left(
    \frac{v^{i,j-1}}{2\Delta y}
    -
    \frac{\nu}{\Delta y^2}
\right)
+
G_u^{i,j+1}
\left(
    -\frac{v^{i,j+1}}{2\Delta y}
    -
    \frac{\nu}{\Delta y^2}
\right)
\nonumber\\
&+
G_v^{i,j}v_x^{i,j}
+
\frac{
    G_{\mathrm{div}}^{i-1,j}
    -
    G_{\mathrm{div}}^{i+1,j}
}{
    2\Delta x
}.
\label{eq:main_u_adjoint}
\end{align}

Equation~\eqref{eq:main_u_adjoint} explicitly represents three types of gradient contributions:
(1) local residual derivatives, (2) neighboring stencil dependencies induced by
finite differences, and (3) cross-field coupling between velocity components and
the divergence constraint. FlashPDE implements
Equation~\eqref{eq:main_u_adjoint} directly in the backward Triton kernel,
avoiding the intermediate derivative graphs required by eager automatic
differentiation.

The backward kernel loads the required forward fields and upstream sensitivities,
evaluates the discrete adjoint terms within the same kernel, and writes the
resulting field gradients. A representative derivation and backward-kernel
pseudocode are provided in
Appendices~\ref{app:derivation} and~\ref{app:algorithms}.

\subsection{Boundary Gradient Correction}
\label{sec:boundary_correction}

Finite-difference stencils near domain boundaries differ from the regular
interior formulation because neighboring points may be unavailable or replaced
by boundary conditions. Directly incorporating all boundary cases into the
interior kernel would introduce additional conditional branches and reduce
execution efficiency.

FlashPDE therefore separates boundary handling from the interior adjoint
computation. The main forward and backward kernels operate on regular interior
tiles, while boundary-specific gradient terms are evaluated through an
independent one-dimensional correction kernel. This design preserves efficient
interior execution while supporting the required boundary-condition treatment
(Appendix~\ref{app:boundary_correction}).

The complete differentiable PDE operator therefore consists of one fused
forward-stencil kernel, one interior analytic-adjoint kernel, and one
boundary-gradient correction kernel. These three launches apply only to the
PDE operator; neural-network, loss, and optimizer operations remain separate.
The resulting operator remains compatible with
\texttt{torch.autograd.Function}, while avoiding the materialization of
individual derivative operations used in eager PyTorch implementations.

\subsection{Numerical Verification}
\label{sec:kernel_verification}

Because the backward operators are derived manually, implementation errors in
indexing or algebraic expressions represent a potential source of numerical
discrepancy. FlashPDE therefore validates each operator against a double
precision PyTorch FD reference implementation using matched physical fields.

For each operator, we measure the scalar residual-loss difference and
input-gradient difference:

\begin{equation*}
    e_{\mathrm{loss}}
    =
    \left|
        \mathcal{L}_{\mathrm{Triton}}
        -
        \mathcal{L}_{\mathrm{PyTorch}}
    \right|,
    \qquad
    e_{\mathrm{grad}}
    =
    \left\|
        \mathbf{g}_{\mathrm{Triton}}
        -
        \mathbf{g}_{\mathrm{PyTorch}}
    \right\|_{\infty}.
\end{equation*}

Across all 14 operators and five deterministic trials per operator, the maximum
observed scalar-loss discrepancy is $2.28\times10^{-7}$, with a corresponding
relative discrepancy of $6.20\times10^{-11}$. The maximum absolute
input-gradient discrepancy is $1.88\times10^{-6}$, and the maximum
infinity-norm relative gradient discrepancy is $4.25\times10^{-8}$. Complete
per-operator results are reported in Appendix~\ref{app:track0}.

These tests show close numerical agreement between FlashPDE operators and
the corresponding PyTorch FD references for the same discrete operators. They do
not evaluate discretization accuracy or PDE solution accuracy, which depend on
the underlying stencil order, grid resolution, boundary treatment, and neural
network optimization.
   
\section{Memory Analysis}
\label{sec:memory}

Coordinate-based PINNs compute higher-order spatial derivatives through repeated
automatic differentiation. Using \texttt{create\_graph=True} keeps the
first-derivative operations differentiable and retains the associated graph
tensors for subsequent derivatives and backward propagation. The resulting
memory cost grows with the number of fields, spatial dimensions, derivative
order, and network activations.

For example, in the profiled three-dimensional heat problem
(\texttt{heat\_3d}), model activations account for 0.82\,GB, whereas constructing
the three second-order spatial derivatives raises peak allocated memory to
16.4\,GB. The complete step-by-step allocation trace is provided in
Appendix~\ref{app:memory_trace}.

Applying finite differences (FD) to the network outputs avoids these nested
coordinate-derivative graphs. The ordinary neural-network autograd graph remains,
but spatial derivatives are evaluated from field values on a Cartesian grid.
The memory columns of Table~\ref{tab:launch_memory} show the resulting
reduction for the six main cases.

Among the completed coordinate-autograd runs, FD reduces peak allocation by
$5.9\times$--$37.0\times$. For the four-field 3D TGV case, coordinate autograd
exceeds 40\,GB, whereas the FD formulation requires 3.39\,GB.

These memory savings result from the change in differentiation strategy and are
shared by eager PyTorch FD (\texttt{eager\_fd}) and FlashPDE. FlashPDE does not
eliminate any additional neural-network graph or introduce a new memory-saving
formulation. Instead, it preserves the low-memory characteristics of the FD
operator while improving its execution efficiency through fused stencil
evaluation and reduced kernel-launch overhead.

Peak memory is measured using
\texttt{torch.cuda.max\_memory\_allocated} over the model forward pass, PDE
evaluation, loss, and backward pass. Detailed measurements for all 17 cases and
the complete protocol are reported in Appendices~\ref{app:memory}
and~\ref{app:memory_protocol}.
\section{Experiments}
\label{sec:experiments}

We evaluate FlashPDE in terms of forward--backward throughput,
time-to-solution (T2S), and kernel-only scaling. All experiments use one
NVIDIA A100-SXM4-40GB GPU with PyTorch 2.9.1 and Triton 3.5.1.
Training uses \texttt{float32}; operator correctness is evaluated separately
in \texttt{float64} in Section~\ref{sec:kernel_verification}.
Complete configurations and measurement protocols are provided in
Appendix~\ref{app:config}.

The six main cases are 1D Burgers, 2D diffusion, two- and three-dimensional
lid-driven cavity (LDC), and two- and three-dimensional Taylor--Green vortex
(TGV). These cases span one to three spatial dimensions, steady and unsteady
equations, and single- to four-field systems~\cite{ghia1982high,taylor1937mechanism,bateman1915some}.
Results for the remaining 11 configurations are reported in
Appendix~\ref{app:track_results}.

We compare four backends:
\texttt{vanilla}, which uses coordinate-based automatic differentiation;
\texttt{eager\_fd}, an eager PyTorch implementation of the output-grid finite-difference residual;
\texttt{compile}, which compiles the neural network while leaving the FD
residual in eager mode; and FlashPDE.
The \texttt{compile} results therefore measure model compilation rather than
full-graph stencil fusion.

\subsection{Forward--Backward Throughput}
\label{sec:full_step}

Table~\ref{tab:microbenchmark} reports the median latency over five runs.
Each run contains 100 warm-up iterations followed by 2,900 measured
iterations. The timed region includes the model forward pass, PDE residual,
loss, and backward pass, but excludes the optimizer update.

\begin{table}[t]
\centering
\caption{
Forward--backward latency and speedup over eager PyTorch FD.
Coordinate autograd is omitted for 3D LDC due to high latency and for 3D TGV
due to GPU OOM.
}
\label{tab:microbenchmark}
\small
\begin{tabular}{lrrrrr}
\toprule
\textbf{Case}
& \textbf{vanilla}
& \textbf{eager FD}
& \textbf{compile}
& \textbf{FlashPDE}
& \textbf{Speedup} \\
\midrule
1D Burgers
& 10.11 & 2.85 & 3.06 & \textbf{2.34} & $1.22\times$ \\

2D Diffusion
& 31.68 & 3.73 & 3.81 & \textbf{3.00} & $1.24\times$ \\

2D LDC
& 18.12 & 6.52 & 6.16 & \textbf{3.84} & $1.70\times$ \\

2D TGV
& 78.53 & 7.61 & 7.59 & \textbf{4.26} & $1.79\times$ \\

3D LDC
& --- & 11.93 & 12.01 & \textbf{5.24} & $\mathbf{2.28\times}$ \\

3D TGV
& --- & 24.40 & \textbf{20.27} & 21.71 & $1.12\times$ \\
\bottomrule
\end{tabular}
\end{table}

FlashPDE improves eager FD latency on all six cases, with speedups from
$1.12\times$ to $2.28\times$. The largest gains occur when eager PDE evaluation
constitutes a substantial fraction of the iteration. The benefit is not
determined by dimension or field count alone: 3D TGV obtains only
$1.12\times$ because more of its execution time remains outside the fused
operator.

Compiling the model has case-dependent effects and is fastest for 3D TGV.
Because the FD residual is not included in the compiled graph, this comparison
does not evaluate TorchInductor's ability to fuse stencil expressions.
A detailed breakdown of acceleration mechanisms is provided in
Appendix~\ref{app:acceleration}.

\subsection{Time-to-Solution}
\label{sec:t2s}

T2S is the wall-clock time at which the PDE loss first crosses a fixed
case-specific threshold, with a maximum budget of 300,000 epochs.
Thresholds are selected in pilot runs and fixed before timing.
For \texttt{eager\_fd} and FlashPDE, we report results over five paired seeds
(42--46); \texttt{vanilla} and \texttt{compile} use one seed.
Compilation, autotuning, optimizer execution, and other cold-start costs are
included in T2S. Complete aggregate results, threshold definitions, epochs, and
effective step times are provided in
Appendices~\ref{app:thresholds} and~\ref{app:main_t2s}.

Although FlashPDE and eager PyTorch FD implement the same discrete stencil,
different floating-point evaluation orders can produce different optimization
trajectories. Operator equivalence is therefore evaluated independently in
Section~\ref{sec:kernel_verification}.

\begin{table*}[t]
\centering
\caption{
Time-to-solution for the six main cases. Results are median $\pm$ standard
deviation over five seeds for \texttt{eager\_fd} and FlashPDE, and single-seed
values for \texttt{compile}. Speedups use the median \texttt{eager\_fd} result,
and case-specific errors are reported as \texttt{eager\_fd}/FlashPDE. Error
definitions and reference solutions are provided in
Appendix~\ref{app:error_metrics}.
}
\label{tab:end2end}
\small
\setlength{\tabcolsep}{5pt}
\begin{tabular}{lrrrrr}
\toprule
\textbf{Case}
& \textbf{eager FD T2S (s)}
& \textbf{compile T2S (s)}
& \textbf{FlashPDE T2S (s)}
& \textbf{Speedup}
& \textbf{Reported error} \\
\midrule
1D Burgers
& $48.5\pm31.1$
& 212.0
& $\mathbf{47.8\pm29.3}$
& $1.01\times$
& 0.51\% / 0.45\% \\

2D Diffusion
& $133.3\pm11.6$
& \textbf{94.8}
& $109.7\pm20.5$
& $1.22\times$
& 2.02\% / 2.07\% \\

2D LDC
& $76.9\pm34.2$
& 591.0
& $\mathbf{43.9\pm10.6}$
& $\mathbf{1.75\times}$
& 4.93\% / 4.66\% \\

2D TGV
& $1026.0\pm144.2$
& 1136.2
& $\mathbf{696.1\pm25.6}$
& $\mathbf{1.47\times}$
& 3.24\% / 3.27\% \\

3D LDC
& $268.4\pm51.9$
& 186.5
& $\mathbf{116.6\pm29.6}$
& $\mathbf{2.30\times}$
& 0.26\% / 0.32\% \\

3D TGV
& $866.7\pm68.5$
& \textbf{454.0}
& $805.0\pm68.8$
& $1.08\times$
& 3.03\% / 3.08\% \\
\bottomrule
\end{tabular}
\end{table*}

FlashPDE has the lowest reported T2S in four cases and improves the median
\texttt{eager\_fd} T2S in all six. The largest improvement is
$2.30\times$ for 3D LDC. The $1.01\times$ result for 1D Burgers is small
relative to its across-seed variation and should be interpreted as parity.

\begin{figure}[t]
\centering
\begin{minipage}[t]{0.48\textwidth}
\centering
\includegraphics[width=\textwidth]{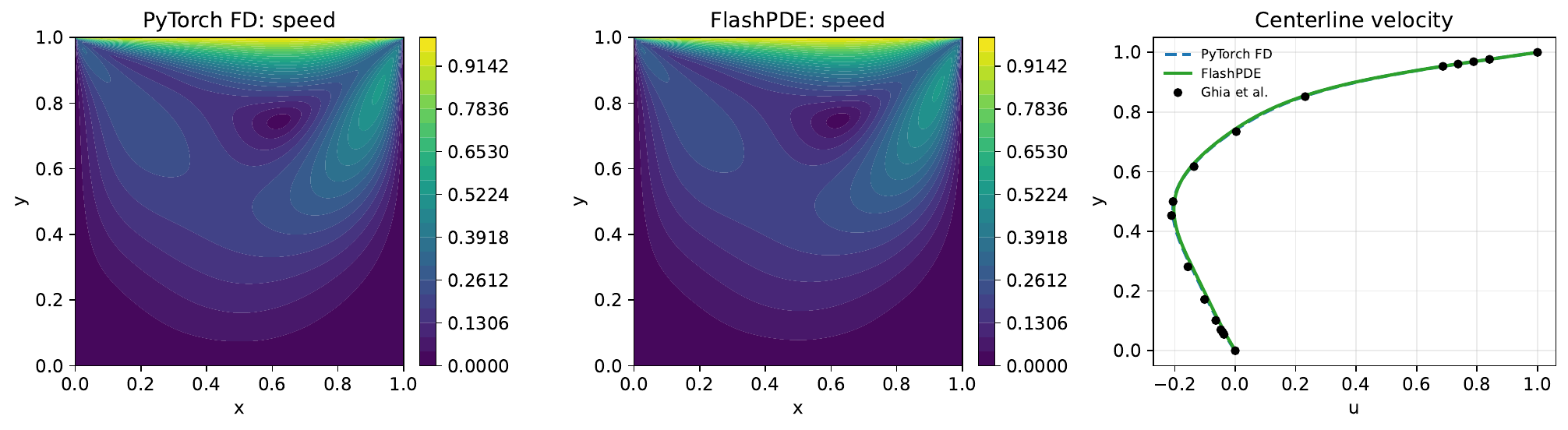}
\end{minipage}
\hfill
\begin{minipage}[t]{0.48\textwidth}
\centering
\includegraphics[width=\textwidth]{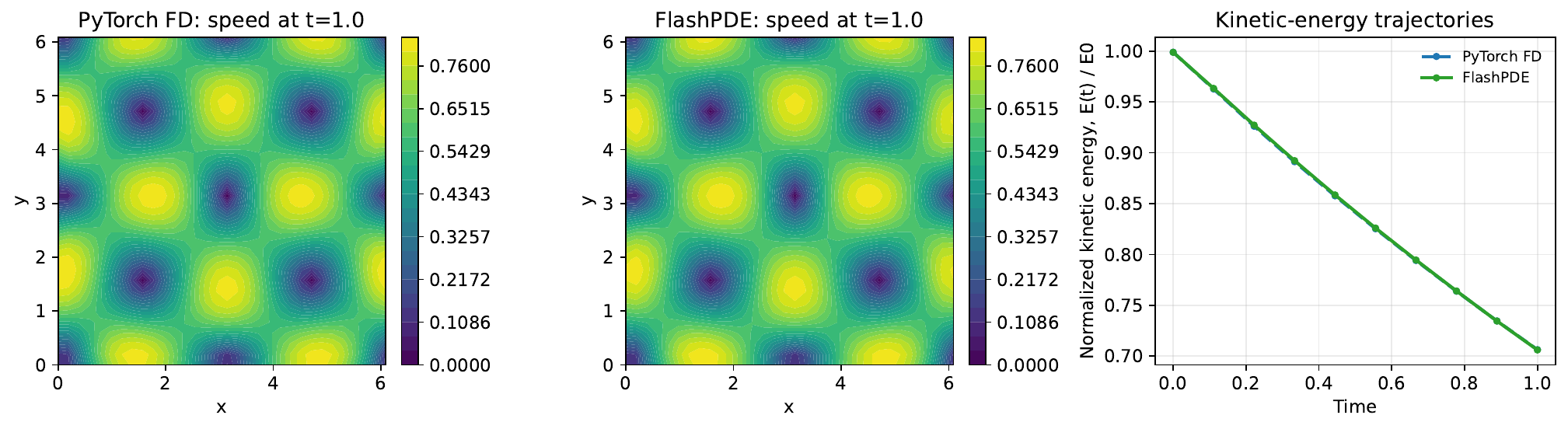}
\end{minipage}
\caption{
Result comparisons for
\textbf{(a)} 2D LDC against Ghia et al.~\cite{ghia1982high} and
\textbf{(b)} PyTorch FD and FlashPDE solution fields and normalized
kinetic-energy trajectories for the 3D TGV configuration.
}
\label{fig:solutions}
\end{figure}

The single-seed compiled-model baseline reports the lowest T2S for 2D diffusion
and 3D TGV. Together with the forward--backward results for 3D TGV, this
indicates that model-side optimization can be important in workloads where
neural-network execution constitutes a large fraction of the training step.
FlashPDE and \texttt{eager\_fd} produce similar median solution errors.
Figure~\ref{fig:solutions} provides two complementary checks: 2D LDC is
compared with the Ghia et al. reference data, while 3D TGV compares the solution
fields and kinetic-energy trajectories produced by eager PyTorch FD and
FlashPDE.

\subsection{Kernel-Only Scaling}
\label{sec:kernel_scaling}

We isolate forward PDE evaluation by benchmarking FlashPDE and eager PyTorch FD
on random field tensors without a neural network or backward pass.
Measurements use \texttt{triton.testing.do\_bench} with 50 warm-up and
100 timed iterations.

\begin{figure*}[t]
\centering
\includegraphics[width=\textwidth]{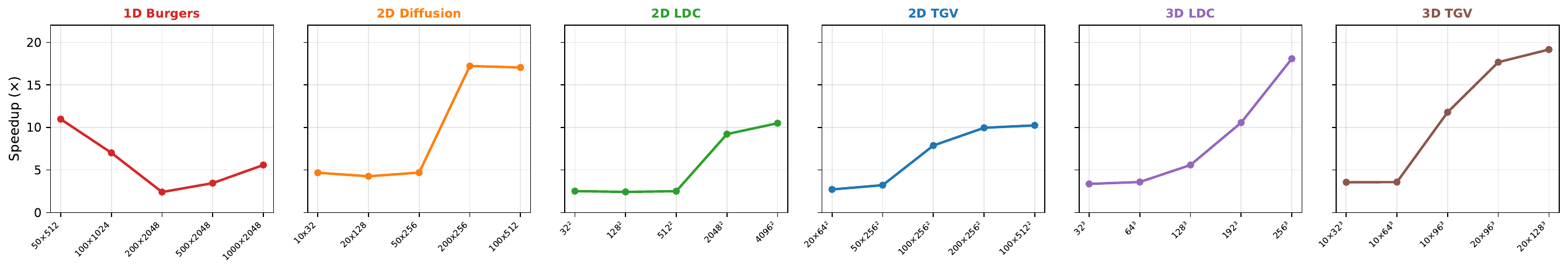}
\caption{
Kernel-only speedup of FlashPDE over eager PyTorch FD, reaching
$19.2\times$ on the largest evaluated 3D TGV grid.
}
\label{fig:scaling}
\end{figure*}

Kernel speedup generally increases for the larger multidimensional problems,
reaching $19.2\times$ for 3D TGV at approximately 42 million space--time
points. The trend is not universal: 1D Burgers remains non-monotonic and has a
smaller large-grid gain. Fusion efficiency therefore depends on stencil
structure, launch geometry, and arithmetic intensity in addition to grid size.

The gap between the maximum kernel-level speedup ($19.2\times$) and
forward--backward speedup ($2.28\times$) reflects the unchanged neural-network
work. The same operators can also be used with compatible CNN field generators;
the corresponding results are provided in Appendix~\ref{app:cnn_track2}.

\section{Related Work}
\label{sec:related}

\subsection{Grid-Based SciML and HPC Solvers}

To alleviate the memory and computational overhead of coordinate-based automatic
differentiation in PINNs~\cite{raissi2019physics,lu2021deepxde}, recent work
has explored grid-based discretizations and numerical-differentiation
formulations. CAN-PINN~\cite{canpinn} combines automatic and numerical
differentiation, while frameworks using RBF-FD
(DTPINN~\cite{sharma2022dtpinn}), hash encodings
(HASH~\cite{huang2024hash}), and tensor formulations
(FastVPINNs~\cite{anandh2024fastvpinn}) improve training efficiency.
However, these methods do not explicitly optimize the GPU execution of
structured-grid stencil operators.

Beyond coordinate-based MLPs, physics-guided convolutional models have been
used to generate physical fields on structured domains. PhyCNN incorporates
governing dynamics into convolutional models for seismic response
prediction~\cite{zhang2019PhyCNN}, Spline-PINN uses Hermite-spline CNN
representations for data-free PDE solving~\cite{wandel2022spline}, and
Physics-informed ConvNet constructs physical fields with a shallow
convolutional architecture~\cite{shi2024convnet}. These approaches primarily
modify the neural field representation. FlashPDE is complementary: it operates
on the resulting grid-aligned field tensors and leaves the neural architecture
unchanged.

Prior work has also optimized high-order stencil computations on
GPUs~\cite{sai2022accelerating}. Traditional structured-grid code-generation
systems, including Devito~\cite{louboutin2019devito},
STELLA~\cite{gysi2015stela}, and OpenSBLI~\cite{jacobs2017opensbli}, generate
optimized stencil programs. GPU-accelerated CFD solvers have also demonstrated
high throughput using specialized CUDA and multi-GPU
implementations~\cite{DBLP:journals/jcphy/WangCP23}. These systems are primarily
designed for standalone numerical simulation rather than neural PDE training
within PyTorch.

Discrete-adjoint systems such as dolfin-adjoint~\cite{farrell2013dolfinadjoint}
automatically derive adjoint models from
high-level finite-element programs. FlashPDE instead targets structured-grid
finite-difference residual operators and implements their discrete adjoints
directly as fused Triton kernels exposed through the PyTorch ecosystem.

\subsection{Hardware-Aware ML Systems}

Triton~\cite{tillet2019triton} has become a widely used framework for
hardware-aware GPU kernel optimization, enabling systems such as
FlashAttention~\cite{dao2022flashattention}, inference engines
(FlashInfer~\cite{ye2024flashinfer}, vLLM~\cite{kwon2023efficient}), and
operator libraries such as FlagGems~\cite{flaggems2024}. However, most
Triton-based optimizations have focused on machine-learning operators, including
attention, matrix operations, and inference workloads, rather than structured
scientific computing kernels.
PyTorch compilation systems optimize and fuse tensor programs~\cite{ansel2024torchcompile},
whereas FlashPDE provides explicitly derived forward and discrete-adjoint
implementations for structured PDE operators.

Differentiable simulation systems such as DiffTaichi~\cite{hu2020difftaichi}
generate efficient gradient kernels for physical simulators. FlashPDE focuses
instead on reusable PDE residual operators embedded
in native PyTorch training workflows. Unlike dense machine-learning operators,
PDE stencils involve spatial neighborhood accesses, multi-field coupling, and
discrete adjoint computations. FlashPDE extends fused-kernel execution
principles to these physics-driven workloads.

\section{Limitations and Conclusion}
\label{sec:conclusion}

FlashPDE currently targets physical fields defined on regular Cartesian grids
and therefore does not directly support random collocation points, irregular
geometries, or unstructured meshes. Each new PDE operator also requires a
manually derived discrete adjoint and a corresponding Triton implementation.
Moreover, FlashPDE preserves the accuracy of the underlying FD discretization
rather than improving it; accuracy remains limited by the stencil, grid
resolution, boundary treatment, and neural-network optimization. Finally,
FlashPDE accelerates only PDE evaluation, so full-step gains are smaller when
network computation dominates. The present evaluation compares against eager PyTorch FD rather than
expert-optimized hand-written CUDA stencil implementations; a direct comparison
with such baselines remains for future work.
The present experiments are limited to one
NVIDIA A100 GPU, and performance portability across other devices and multi-GPU
systems remains to be evaluated.

We presented \textbf{FlashPDE}, a PyTorch-integrated Triton library for fused
structured-grid PDE operators. Its custom \texttt{torch.autograd.Function}
implementations combine a fused forward stencil, a discrete analytic-adjoint
backward pass, and boundary-gradient correction. The current library contains 14 architecture-independent, drop-in
differentiable PDE operators covering 17 PDE configurations. Because these
operators consume grid-aligned physical field tensors rather than network
internals, the same implementations can be reused across compatible MLP and
CNN field generators.

Across six representative cases, FlashPDE achieves
$1.12\times$--$2.28\times$ forward--backward speedups and up to
$2.30\times$ median time-to-solution speedup over eager PyTorch FD.
Kernel-only speedup reaches $19.2\times$ on the largest evaluated grid.
Output-grid FD reduces the 3D TGV memory footprint from more than $40$\,GB to
3.4\,GB, while FlashPDE preserves this low-memory formulation and reduces the
execution overhead of the resulting FD operators.

Future work includes automatic generation of discrete adjoints and Triton
kernels, support for higher-order and irregular-grid discretizations, evaluation
on additional GPU architectures, and integration with full-graph compilation.
The implementation is available at
\url{https://github.com/factnn/FlashPDE}.


\clearpage
\bibliographystyle{unsrtnat} 
\bibliography{main} 

\newpage
\appendix
\appendix

\section{Operator Design Details}
\label{app:full_results}

\subsection{Operator Coverage}
\label{app:operator_coverage}

FlashPDE contains 14 differentiable PDE operators covering 17 benchmark configurations.
Several operators are reused across PDE configurations that share the same
stencil structure but differ in coefficients, initial conditions, or forcing
terms.
The extended set includes the Sod shock-tube and Allen--Cahn
benchmarks~\cite{sod1978survey,allen1979microscopic}.

\begin{table*}[t]
\centering
\caption{
FlashPDE operator coverage. Multiple benchmark configurations may share one operator implementation
when their discrete residual structure is identical.
}
\label{tab:operator_coverage}
\small
\begin{tabular}{llll}
\toprule
\textbf{Operator ID}
& \textbf{Domain}
& \textbf{Benchmark configuration(s)}
& \textbf{Equation class} \\
\midrule
\texttt{1d\_steady}
& 1D
& \texttt{burgers\_1d\_steady}
& Steady Burgers \\

\texttt{1d\_unsteady}
& 1D+T
& \texttt{burgers\_1d\_unsteady}
& Unsteady Burgers \\

\texttt{1d\_heat}
& 1D+T
& \texttt{diffusion\_1d}
& Diffusion / heat \\

\texttt{2d\_compressible}
& 1D+T
& \texttt{sod\_1d}
& Compressible Euler \\

\texttt{2d\_poisson}
& 2D
& \texttt{poisson\_2d}
& Elliptic \\

\texttt{2d\_transport}
& 2D+T
& \texttt{transport\_2d}, \texttt{diffusion\_2d}, \texttt{heat\_2d}
& Transport / diffusion \\

\texttt{2d\_wave}
& 2D+T
& \texttt{wave\_2d}
& Hyperbolic wave \\

\texttt{2d\_allencahn}
& 1D+T
& \texttt{allen\_cahn}
& Reaction--diffusion \\

\texttt{2d\_ns\_steady}
& 2D
& \texttt{ldc\_2d}
& Steady Navier--Stokes \\

\texttt{2d\_ns\_unsteady}
& 2D+T
& \texttt{tgv\_2d}
& Unsteady Navier--Stokes \\

\texttt{3d\_ns\_steady}
& 3D
& \texttt{ldc\_3d}
& Steady Navier--Stokes \\

\texttt{3d\_ns\_unsteady}
& 3D+T
& \texttt{tgv\_3d}, \texttt{tgv\_3d\_smooth}
& Unsteady Navier--Stokes \\

\texttt{3d\_poisson}
& 3D
& \texttt{poisson\_3d}
& Elliptic \\

\texttt{3d\_heat}
& 3D+T
& \texttt{heat\_3d}
& Diffusion / heat \\
\bottomrule
\end{tabular}
\end{table*}

\subsection{Eager PyTorch and Fused Triton Implementations}
\label{app:code_comparison}

Figure~\ref{fig:code_comparison_appendix} illustrates how an eager PyTorch
finite-difference residual is mapped to a fused Triton forward kernel. Tensor
slices are commonly views, but the arithmetic operations that combine them
produce separate kernels and derivative intermediates. FlashPDE evaluates the
same stencil inside one forward kernel and writes only the residual fields.

\begin{figure*}[t]
\centering

\begin{minipage}[t]{0.48\textwidth}
\centering
\textbf{Eager PyTorch FD}\\[3pt]
\begin{tcolorbox}[
    colback=gray!5,
    colframe=gray!60,
    fontupper=\ttfamily\scriptsize,
    left=2pt,
    right=2pt,
    top=2pt,
    bottom=2pt,
    boxrule=0.5pt
]
\# Tensor slices followed by eager arithmetic\\
u\_x = (U[2:,1:-1] - U[:-2,1:-1]) / (2*dx)\\
u\_y = (U[1:-1,2:] - U[1:-1,:-2]) / (2*dy)\\
u\_xx = (U[2:,1:-1] - 2*U[1:-1,1:-1]\\
\hspace*{1.1cm} + U[:-2,1:-1]) / dx**2\\
u\_yy = (U[1:-1,2:] - 2*U[1:-1,1:-1]\\
\hspace*{1.1cm} + U[1:-1,:-2]) / dy**2\\
v\_x = (V[2:,1:-1] - V[:-2,1:-1]) / (2*dx)\\
\textit{[... additional derivatives ...]}\\
res\_u = uc*u\_x + vc*u\_y + p\_x\\
\hspace*{1.1cm} - nu*(u\_xx + u\_yy)\\
res\_v = uc*v\_x + vc*v\_y + p\_y\\
\hspace*{1.1cm} - nu*(v\_xx + v\_yy)\\
res\_div = u\_x + v\_y\\[3pt]
\# Multiple launches and intermediate tensors
\end{tcolorbox}
\end{minipage}
\hfill
\begin{minipage}[t]{0.48\textwidth}
\centering
\textbf{FlashPDE Fused Forward Kernel}\\[3pt]
\begin{tcolorbox}[
    colback=blue!3,
    colframe=blue!40,
    fontupper=\ttfamily\scriptsize,
    left=2pt,
    right=2pt,
    top=2pt,
    bottom=2pt,
    boxrule=0.5pt
]
\# One Triton program processes one tile\\
@triton.jit\\
def ldc\_fwd\_kernel(U, V, P, resU, ...):\\
~~pid\_x = program\_id(0)\\
~~pid\_y = program\_id(1)\\
~~\# Load the local field stencil\\
~~u\_c  = load(U[ix, iy])\\
~~u\_xp = load(U[ix+1, iy])\\
~~u\_xm = load(U[ix-1, iy])\\
~~\textit{[... remaining field values ...]}\\
~~\# Evaluate derivatives on chip\\
~~u\_x  = (u\_xp - u\_xm) / (2*dx)\\
~~u\_xx = (u\_xp - 2*u\_c + u\_xm) / dx**2\\
~~\textit{[... remaining derivatives ...]}\\
~~\# Evaluate and store the final residuals\\
~~res\_u = u\_c*u\_x + v\_c*u\_y + p\_x\\
~~~~~~~~~~- nu*(u\_xx + u\_yy)\\
~~\textit{[... res\_v and res\_div ...]}\\
~~store(resU, resV, resDiv)\\[3pt]
\# One launch for the complete forward residual
\end{tcolorbox}
\end{minipage}

\caption{
Representative eager PyTorch FD and fused FlashPDE forward implementations.
FlashPDE evaluates the derivative and residual expressions in one Triton launch
instead of materializing the individual derivative tensors.
}
\label{fig:code_comparison_appendix}
\end{figure*}

\subsection{Discrete Analytic-Adjoint Derivation for the
\texorpdfstring{$u$}{u}-Velocity Field}
\label{app:derivation}

This section derives the horizontal-velocity component of the discrete adjoint
for the steady two-dimensional incompressible Navier--Stokes operator. At each
interior grid point $(i,j)$, the residuals are
\begin{align*}
    \mathrm{res}_u^{i,j}
    &=
    u^{i,j}u_x^{i,j}
    +
    v^{i,j}u_y^{i,j}
    +
    p_x^{i,j}
    -
    \nu\left(
        u_{xx}^{i,j}
        +
        u_{yy}^{i,j}
    \right),
    \\
    \mathrm{res}_v^{i,j}
    &=
    u^{i,j}v_x^{i,j}
    +
    v^{i,j}v_y^{i,j}
    +
    p_y^{i,j}
    -
    \nu\left(
        v_{xx}^{i,j}
        +
        v_{yy}^{i,j}
    \right),
    \\
    \mathrm{res}_{\mathrm{div}}^{i,j}
    &=
    u_x^{i,j}
    +
    v_y^{i,j}.
\end{align*}

For a scalar field $q$, the centered differences are
\begin{align*}
    q_x^{i,j}
    &=
    \frac{q^{i+1,j}-q^{i-1,j}}{2\Delta x},
    &
    q_{xx}^{i,j}
    &=
    \frac{q^{i+1,j}-2q^{i,j}+q^{i-1,j}}{\Delta x^2},
    \\
    q_y^{i,j}
    &=
    \frac{q^{i,j+1}-q^{i,j-1}}{2\Delta y},
    &
    q_{yy}^{i,j}
    &=
    \frac{q^{i,j+1}-2q^{i,j}+q^{i,j-1}}{\Delta y^2}.
\end{align*}

The custom \texttt{autograd.Function} receives
\begin{equation*}
    G_u
    =
    \frac{\partial\mathcal{L}}{\partial\mathrm{res}_u},
    \qquad
    G_v
    =
    \frac{\partial\mathcal{L}}{\partial\mathrm{res}_v},
    \qquad
    G_{\mathrm{div}}
    =
    \frac{\partial\mathcal{L}}
         {\partial\mathrm{res}_{\mathrm{div}}}.
\end{equation*}

A perturbation of $u^{i,j}$ affects the local $u$-momentum residual, four
neighboring $u$-momentum residuals, the local $v$-momentum residual, and two
neighboring continuity residuals:
\begin{align}
\frac{\partial\mathcal{L}}{\partial u^{i,j}}
={}&
G_u^{i,j}
\frac{\partial\mathrm{res}_u^{i,j}}{\partial u^{i,j}}
+
G_u^{i-1,j}
\frac{\partial\mathrm{res}_u^{i-1,j}}{\partial u^{i,j}}
+
G_u^{i+1,j}
\frac{\partial\mathrm{res}_u^{i+1,j}}{\partial u^{i,j}}
\nonumber\\
&+
G_u^{i,j-1}
\frac{\partial\mathrm{res}_u^{i,j-1}}{\partial u^{i,j}}
+
G_u^{i,j+1}
\frac{\partial\mathrm{res}_u^{i,j+1}}{\partial u^{i,j}}
+
G_v^{i,j}
\frac{\partial\mathrm{res}_v^{i,j}}{\partial u^{i,j}}
\nonumber\\
&+
G_{\mathrm{div}}^{i-1,j}
\frac{
    \partial\mathrm{res}_{\mathrm{div}}^{i-1,j}
}{
    \partial u^{i,j}
}
+
G_{\mathrm{div}}^{i+1,j}
\frac{
    \partial\mathrm{res}_{\mathrm{div}}^{i+1,j}
}{
    \partial u^{i,j}
}.
\label{eq:app_adjoint_summation}
\end{align}

The local contribution is
\begin{equation*}
    \frac{
        \partial\mathrm{res}_u^{i,j}
    }{
        \partial u^{i,j}
    }
    =
    u_x^{i,j}
    +
    \frac{2\nu}{\Delta x^2}
    +
    \frac{2\nu}{\Delta y^2}.
\end{equation*}

The horizontal-neighbor contributions are
\begin{align*}
    \frac{
        \partial\mathrm{res}_u^{i-1,j}
    }{
        \partial u^{i,j}
    }
    &=
    \frac{u^{i-1,j}}{2\Delta x}
    -
    \frac{\nu}{\Delta x^2},
    \\
    \frac{
        \partial\mathrm{res}_u^{i+1,j}
    }{
        \partial u^{i,j}
    }
    &=
    -
    \frac{u^{i+1,j}}{2\Delta x}
    -
    \frac{\nu}{\Delta x^2}.
\end{align*}

The vertical-neighbor contributions are
\begin{align*}
    \frac{
        \partial\mathrm{res}_u^{i,j-1}
    }{
        \partial u^{i,j}
    }
    &=
    \frac{v^{i,j-1}}{2\Delta y}
    -
    \frac{\nu}{\Delta y^2},
    \\
    \frac{
        \partial\mathrm{res}_u^{i,j+1}
    }{
        \partial u^{i,j}
    }
    &=
    -
    \frac{v^{i,j+1}}{2\Delta y}
    -
    \frac{\nu}{\Delta y^2}.
\end{align*}

The cross-equation contributions are
\begin{align*}
    \frac{
        \partial\mathrm{res}_v^{i,j}
    }{
        \partial u^{i,j}
    }
    &=
    v_x^{i,j},
    \\
    \frac{
        \partial\mathrm{res}_{\mathrm{div}}^{i-1,j}
    }{
        \partial u^{i,j}
    }
    &=
    \frac{1}{2\Delta x},
    \\
    \frac{
        \partial\mathrm{res}_{\mathrm{div}}^{i+1,j}
    }{
        \partial u^{i,j}
    }
    &=
    -
    \frac{1}{2\Delta x}.
\end{align*}

Substitution into Equation~\eqref{eq:app_adjoint_summation} gives
\begin{align}
\frac{\partial\mathcal{L}}{\partial u^{i,j}}
={}&
G_u^{i,j}
\left(
    u_x^{i,j}
    +
    \frac{2\nu}{\Delta x^2}
    +
    \frac{2\nu}{\Delta y^2}
\right)
\nonumber\\
&+
G_u^{i-1,j}
\left(
    \frac{u^{i-1,j}}{2\Delta x}
    -
    \frac{\nu}{\Delta x^2}
\right)
+
G_u^{i+1,j}
\left(
    -
    \frac{u^{i+1,j}}{2\Delta x}
    -
    \frac{\nu}{\Delta x^2}
\right)
\nonumber\\
&+
G_u^{i,j-1}
\left(
    \frac{v^{i,j-1}}{2\Delta y}
    -
    \frac{\nu}{\Delta y^2}
\right)
+
G_u^{i,j+1}
\left(
    -
    \frac{v^{i,j+1}}{2\Delta y}
    -
    \frac{\nu}{\Delta y^2}
\right)
\nonumber\\
&+
G_v^{i,j}v_x^{i,j}
+
\frac{
    G_{\mathrm{div}}^{i-1,j}
    -
    G_{\mathrm{div}}^{i+1,j}
}{
    2\Delta x
}.
\label{eq:app_final_u_adjoint}
\end{align}

FlashPDE implements Equation~\eqref{eq:app_final_u_adjoint} directly in the
backward Triton kernel. The remaining field gradients are derived in the same
way from the transpose of the corresponding discrete residual Jacobian.

\subsection{Representative Forward and Backward Kernel Pseudocode}
\label{app:algorithms}

Figure~\ref{fig:pseudocode} gives representative pseudocode for the
two-dimensional LDC operator. The backward panel expands the $u$-field
component; the implemented kernel also evaluates the $v$- and $p$-field
gradients.

\begin{figure*}[t]
\centering

\begin{minipage}[t]{0.48\textwidth}
\centering
\textbf{Forward kernel: \texttt{ldc\_2d}}\\[3pt]
\begin{tcolorbox}[
    colback=gray!5,
    colframe=gray!60,
    fontupper=\ttfamily\scriptsize,
    left=2pt,
    right=2pt,
    top=2pt,
    bottom=2pt,
    boxrule=0.5pt
]
Input: U, V, P, dx, dy, nu\\
Output: res\_u, res\_v, res\_div\\[3pt]

px, py = program\_id(0), program\_id(1)\\
for (i,j) in interior tile:\\
~~load u\_c, u\_xp, u\_xm, u\_yp, u\_ym\\
~~load v\_c, v\_xp, v\_xm, v\_yp, v\_ym\\
~~load p\_xp, p\_xm, p\_yp, p\_ym\\[2pt]

~~u\_x  = (u\_xp - u\_xm) / (2*dx)\\
~~u\_y  = (u\_yp - u\_ym) / (2*dy)\\
~~u\_xx = (u\_xp - 2*u\_c + u\_xm) / dx**2\\
~~u\_yy = (u\_yp - 2*u\_c + u\_ym) / dy**2\\
~~\textit{[evaluate v and p derivatives]}\\[2pt]

~~res\_u = u\_c*u\_x + v\_c*u\_y + p\_x\\
~~~~~~~~~~- nu*(u\_xx + u\_yy)\\
~~res\_v = u\_c*v\_x + v\_c*v\_y + p\_y\\
~~~~~~~~~~- nu*(v\_xx + v\_yy)\\
~~res\_div = u\_x + v\_y\\[2pt]

~~store(res\_u, res\_v, res\_div)
\end{tcolorbox}
\end{minipage}
\hfill
\begin{minipage}[t]{0.48\textwidth}
\centering
\textbf{Backward kernel: $u$-field component}\\[3pt]
\begin{tcolorbox}[
    colback=blue!3,
    colframe=blue!40,
    fontupper=\ttfamily\scriptsize,
    left=2pt,
    right=2pt,
    top=2pt,
    bottom=2pt,
    boxrule=0.5pt
]
Input: U, V, G\_u, G\_v, G\_div, grid params\\
Output: grad\_U, grad\_V, grad\_P\\[3pt]

idx  = 1 / (2*dx); idx2 = 1 / dx**2\\
idy  = 1 / (2*dy); idy2 = 1 / dy**2\\[2pt]

for (i,j) in interior tile:\\
~~load required U and V stencil values\\
~~load neighboring upstream sensitivities\\[2pt]

~~u\_x = (u\_xp - u\_xm) * idx\\
~~v\_x = (v\_xp - v\_xm) * idx\\[2pt]

~~grad\_u = g\_u\_c * (u\_x + 2*nu*idx2\\
~~~~~~~~~~~~~~~~~~~~~~+ 2*nu*idy2)\\
~~grad\_u += g\_u\_xm * (u\_xm*idx - nu*idx2)\\
~~grad\_u += g\_u\_xp * (-u\_xp*idx - nu*idx2)\\
~~grad\_u += g\_u\_ym * (v\_ym*idy - nu*idy2)\\
~~grad\_u += g\_u\_yp * (-v\_yp*idy - nu*idy2)\\
~~grad\_u += g\_v\_c * v\_x\\
~~grad\_u += (g\_div\_xm - g\_div\_xp) * idx\\[2pt]

~~\textit{[evaluate grad\_v and grad\_p]}\\
~~store(grad\_u, grad\_v, grad\_p)
\end{tcolorbox}
\end{minipage}

\caption{
Representative pseudocode for the \texttt{ldc\_2d} fused forward and backward
kernels. Other operators follow the same pattern but differ in field count,
stencil geometry, residual definition, and boundary treatment.
}
\label{fig:pseudocode}
\end{figure*}

\subsection{Boundary-Gradient Correction}
\label{app:boundary_correction}

The fused backward kernel evaluates the regular interior transpose stencil.
Boundary points require different contributions because prescribed values,
one-sided formulas, or omitted neighbors change the discrete dependency graph.
FlashPDE applies these terms in a separate one-dimensional correction kernel
rather than branching over boundary cases inside the interior kernel.

For the current operators, one differentiable PDE evaluation therefore consists
of one forward-stencil launch, one interior analytic-adjoint launch, and one
boundary-gradient correction launch. This count applies only to the PDE
operator; model, loss, and optimizer kernels remain separate.

\subsection{Triton Autotuning}
\label{app:autotune}

Each FlashPDE kernel uses \texttt{@triton.autotune} to select a launch
configuration from a predefined candidate set. On the first invocation for a
new key, Triton benchmarks the supplied configurations, caches the fastest
observed candidate, and reuses it in subsequent calls.

For example, the two-dimensional steady Navier--Stokes forward kernel uses
\begin{verbatim}
@triton.autotune(
    configs=[
        triton.Config({'BLOCK_X': 8,  'BLOCK_Y': 8}),
        triton.Config({'BLOCK_X': 16, 'BLOCK_Y': 16}),
        triton.Config({'BLOCK_X': 32, 'BLOCK_Y': 32}),
        triton.Config({'BLOCK_X': 16, 'BLOCK_Y': 32}),
    ],
    key=['Nx', 'Ny'],
)
\end{verbatim}

One-dimensional kernels use linear block sizes between 64 and 512. The current
three-dimensional kernels flatten the spatial domain and evaluate block sizes up
to 512. Autotuning selects the best configuration among the supplied candidates;
it does not guarantee a globally optimal configuration.

\subsection{Numerical Verification for All Operators}
\label{app:track0}

Every operator is compared with a double-precision PyTorch FD reference in five
deterministic trials. Input fields are sampled from a standard normal distribution
with seeds 42--46; the compressible case additionally transforms density and energy
to physically admissible positive values. The scalar residual losses are
differentiated with respect to every physical input field. The table reports the
absolute scalar-loss discrepancy $e_{\mathrm{loss}}$ and the maximum absolute
input-gradient discrepancy $e_{\mathrm{grad}}$ over all grid entries and physical
fields and five trials. The table also reports the maximum relative gradient
discrepancy $e_{\mathrm{grad,rel}}=e_{\mathrm{grad}}/
\|\mathbf{g}_{\mathrm{PyTorch}}\|_\infty$.

\begin{table*}[t]
\centering
\caption{
Numerical verification for all 14 FlashPDE operators against the
double-precision PyTorch FD reference using five deterministic float64 trials
per operator (standard-normal inputs, seeds 42--46). Each entry is the maximum
over the five trials.
}
\label{tab:verification_all}
\small
\begin{tabular}{lccc}
\toprule
\textbf{Operator ID}
& \textbf{$e_{\mathrm{loss}}$}
& \textbf{$e_{\mathrm{grad}}$}
& \textbf{$e_{\mathrm{grad,rel}}$} \\
\midrule
\texttt{1d\_steady}
& $1.819\times10^{-12}$ & $2.274\times10^{-13}$ & $2.415\times10^{-16}$ \\
\texttt{1d\_unsteady}
& $2.842\times10^{-14}$ & $3.880\times10^{-8}$ & $1.186\times10^{-9}$ \\
\texttt{1d\_heat}
& $0$ & $9.095\times10^{-13}$ & $2.811\times10^{-16}$ \\
\texttt{2d\_compressible}
& $2.283\times10^{-7}$ & $7.171\times10^{-8}$ & $1.181\times10^{-10}$ \\
\texttt{2d\_poisson}
& $0$ & $7.276\times10^{-12}$ & $2.054\times10^{-16}$ \\
\texttt{2d\_transport}
& $0$ & $4.466\times10^{-9}$ & $1.123\times10^{-8}$ \\
\texttt{2d\_wave}
& $4.547\times10^{-13}$ & $1.881\times10^{-6}$ & $4.251\times10^{-8}$ \\
\texttt{2d\_allencahn}
& $7.105\times10^{-15}$ & $1.051\times10^{-7}$ & $8.618\times10^{-9}$ \\
\texttt{2d\_ns\_steady}
& $2.274\times10^{-13}$ & $1.421\times10^{-14}$ & $2.894\times10^{-16}$ \\
\texttt{2d\_ns\_unsteady}
& $1.421\times10^{-14}$ & $4.110\times10^{-9}$ & $7.740\times10^{-9}$ \\
\texttt{3d\_ns\_steady}
& $1.705\times10^{-13}$ & $9.538\times10^{-8}$ & $7.942\times10^{-9}$ \\
\texttt{3d\_ns\_unsteady}
& $7.105\times10^{-15}$ & $8.800\times10^{-9}$ & $3.094\times10^{-8}$ \\
\texttt{3d\_poisson}
& $4.366\times10^{-11}$ & $1.364\times10^{-12}$ & $4.462\times10^{-16}$ \\
\texttt{3d\_heat}
& $1.776\times10^{-15}$ & $6.033\times10^{-10}$ & $4.315\times10^{-9}$ \\
\bottomrule
\end{tabular}
\end{table*}

\subsection{Profiler Breakdown}
\label{app:profiler}

Table~\ref{tab:profiler_all} reports CUDA launch counts for one forward--loss--backward
iteration, including the model forward pass, PDE residual, boundary-condition
loss, and backward pass, but excluding the optimizer update. Counts are obtained from
\texttt{torch.profiler} traces by counting GPU kernel activity records within the measured forward--loss--backward region. CUDA
memory-copy and memory-set activity records are not counted as kernel launches.

\begin{table}[t]
\centering
\caption{CUDA kernel launches per forward--loss--backward iteration for all 17 cases.}
\label{tab:profiler_all}
\small
\begin{tabular}{lrrc}
\toprule
\textbf{Case}
& \textbf{PyTorch FD}
& \textbf{FlashPDE}
& \textbf{Reduction} \\
\midrule
\texttt{burgers\_1d\_steady}   & 94  & 70  & $1.3\times$ \\
\texttt{burgers\_1d\_unsteady} & 129 & 76  & $1.7\times$ \\
\texttt{diffusion\_1d}         & 107 & 86  & $1.2\times$ \\
\texttt{sod\_1d}               & 335 & 309 & $1.1\times$ \\
\texttt{ldc\_2d}               & 321 & 137 & $2.3\times$ \\
\texttt{tgv\_2d}               & 452 & 182 & $2.5\times$ \\
\texttt{poisson\_2d}           & 115 & 76  & $1.5\times$ \\
\texttt{transport\_2d}         & 192 & 84  & $2.3\times$ \\
\texttt{diffusion\_2d}         & 192 & 84  & $2.3\times$ \\
\texttt{heat\_2d}              & 154 & 84  & $1.8\times$ \\
\texttt{wave\_2d}              & 172 & 113 & $1.5\times$ \\
\texttt{allen\_cahn}           & 125 & 94  & $1.3\times$ \\
\texttt{ldc\_3d}               & 727 & 231 & $3.1\times$ \\
\texttt{tgv\_3d}               & 934 & 267 & $3.5\times$ \\
\texttt{tgv\_3d\_smooth}       & 934 & 267 & $3.5\times$ \\
\texttt{poisson\_3d}           & 170 & 93  & $1.8\times$ \\
\texttt{heat\_3d}              & 216 & 127 & $1.7\times$ \\
\bottomrule
\end{tabular}
\end{table}

\subsection{Sources of Operator-Level Acceleration}
\label{app:acceleration}

\subsubsection{Triton Implementation Rationale}
\label{app:triton_rationale}

FlashPDE uses Triton rather than handwritten CUDA extensions to provide
differentiable fused PDE operators within the PyTorch ecosystem. A specialized
CUDA implementation can potentially achieve higher peak performance for a fixed
kernel, but requires additional integration effort for custom autograd
operators, memory management, and architecture-specific optimization. Triton
provides a programmable GPU kernel abstraction with direct integration into
PyTorch workflows, making it suitable for developing a reusable operator
library across multiple PDE configurations. The goal of FlashPDE is therefore
not to replace highly optimized application-specific CUDA solvers, but to
provide a practical execution layer that combines GPU kernel fusion with
differentiable PDE training.

FlashPDE's operator-level speedup comes from four related mechanisms:
\begin{enumerate}
    \item \textbf{Fewer kernel launches.}
    Eager FD decomposes a residual into multiple arithmetic and reduction
    kernels. FlashPDE uses one forward launch and two additional launches for
    analytic backpropagation and boundary correction.

    \item \textbf{Reduced global-memory traffic.}
    Quantities such as $u_x$, $u_{xx}$, and $v_y$ are consumed inside the fused
    operator instead of being written to and read from global-memory
    intermediates.

    \item \textbf{Fewer temporary tensor allocations.}
    FlashPDE still allocates residual outputs, saved tensors, and field
    gradients, but avoids many short-lived derivative tensors created by the
    eager expression.

    \item \textbf{Autotuned launch configurations.}
    Triton benchmarks a finite set of block configurations and caches the
    fastest observed candidate for the current key and device.
\end{enumerate}

The contribution of each mechanism depends on field count, stencil width, tensor
geometry, and the fraction of the full training step spent in PDE evaluation.

\section{Memory Attribution Details}
\label{app:memory_details}

\subsection{Step-by-Step Autograd Memory Trace}
\label{app:memory_trace}

We profile the three-dimensional unsteady heat equation
(\texttt{heat\_3d}) on an NVIDIA A100-SXM4-40GB GPU. The configuration contains
$10\times32^3$ space--time points and uses a five-layer MLP with hidden width
128. Table~\ref{tab:memory_trace} reports the allocated memory observed after each
construction stage. The peak of 16.38\,GB occurs at the third second-order
derivative; after \texttt{backward()}, retained graph tensors are released and
allocation drops to 0.21\,GB. The full forward--loss--backward peak for
\texttt{heat\_3d} is 17.83\,GB (Table~\ref{tab:memory_all}), which is higher
because it includes loss evaluation and backward propagation in addition to
residual construction.

\begin{table}[t]
\centering
\caption{
Allocated GPU memory observed after each residual-construction stage for
\texttt{heat\_3d}. The peak is reached at step~6; after
\texttt{backward()}, graph tensors are released.
}
\label{tab:memory_trace}
\small
\begin{tabular}{clr}
\toprule
\textbf{Step} & \textbf{Operation} & \textbf{Allocated (GB)} \\
\midrule
1 & Input coordinates & 0.005 \\
2 & Model activations & 0.820 \\
3 & First-order derivatives & 2.430 \\
4 & $\partial^2/\partial x^2$ (\texttt{create\_graph=True}) & 7.080 \\
5 & $\partial^2/\partial y^2$ (\texttt{create\_graph=True}) & 11.730 \\
6 & $\partial^2/\partial z^2$ (\texttt{create\_graph=True}) & 16.380 \\
7 & PDE residual and loss & 16.400 \\
8 & After \texttt{backward()} & 0.210 \\
\bottomrule
\end{tabular}
\end{table}

The model activations raise the allocated memory to 0.82\,GB. Constructing the three
second-order spatial derivatives raises it to 16.38\,GB, with an increase of
approximately 4.65\,GB per additional derivative in this configuration. About
95\% of the observed peak relative to the coordinate and model allocations is
associated with retained derivative graphs and their intermediates. This
percentage is specific to the profiled setup and is not assumed to hold for all
PDEs or network architectures.

\subsection{Memory Results for All Cases}
\label{app:memory}

Table~\ref{tab:memory_all} reports peak allocated memory under coordinate-based
automatic differentiation and output-grid FD. Eager PyTorch FD and FlashPDE use
the same differentiation strategy and have the same reported peak allocation to
the displayed precision.

\begin{table}[t]
\centering
\caption{
Peak PyTorch-allocated GPU memory across all 17 configurations.
OOM indicates that coordinate autograd exceeds the 40\,GB device capacity.
}
\label{tab:memory_all}
\small
\begin{tabular}{lrrrc}
\toprule
\textbf{Case}
& \textbf{Grid}
& \textbf{Autograd}
& \textbf{FD}
& \textbf{Reduction} \\
&
&
\textbf{(GB)}
& \textbf{(GB)}
& \\
\midrule
\texttt{burgers\_1d\_steady}
& 256 & 0.021 & 0.019 & $1.1\times$ \\
\texttt{burgers\_1d\_unsteady}
& $1024\times100$ & 1.075 & 0.183 & $5.9\times$ \\
\texttt{diffusion\_1d}
& $128\times50$ & 0.083 & 0.028 & $3.0\times$ \\
\texttt{sod\_1d}
& $1000\times200$ & 11.57 & 1.26 & $9.2\times$ \\
\texttt{ldc\_2d}
& $128^2$ & 1.321 & 0.086 & $15.4\times$ \\
\texttt{tgv\_2d}
& $64^2\times20$ & 7.830 & 0.399 & $19.6\times$ \\
\texttt{poisson\_2d}
& $64^2$ & 0.179 & 0.036 & $5.0\times$ \\
\texttt{transport\_2d}
& $64^2\times20$ & 3.254 & 0.356 & $9.1\times$ \\
\texttt{diffusion\_2d}
& $64^2\times20$ & 3.254 & 0.356 & $9.1\times$ \\
\texttt{heat\_2d}
& $64^2\times20$ & 3.253 & 0.356 & $9.1\times$ \\
\texttt{wave\_2d}
& $64^2\times20$ & 4.428 & 0.356 & $12.4\times$ \\
\texttt{allen\_cahn}
& $100\times256$ & 1.100 & 0.220 & $5.0\times$ \\
\texttt{ldc\_3d}
& $48^3$ & 17.59 & 0.47 & $37.0\times$ \\
\texttt{tgv\_3d}
& $32^3\times10$ & OOM & 3.39 & --- \\
\texttt{tgv\_3d\_smooth}
& $32^3\times10$ & OOM & 3.39 & --- \\
\texttt{poisson\_3d}
& $32^3$ & 1.783 & 0.154 & $11.6\times$ \\
\texttt{heat\_3d}
& $32^3\times10$ & 17.83 & 1.200 & $14.9\times$ \\
\bottomrule
\end{tabular}
\end{table}

\subsection{Memory Measurement Protocol}
\label{app:memory_protocol}

Peak active allocation is measured using
\texttt{torch.cuda.max\_memory\_allocated}. Before each measurement,
\texttt{torch.cuda.reset\_peak\_memory\_stats} clears the recorded peak.
The measured region includes model forward evaluation, PDE residual
construction, loss evaluation, and backward propagation, and ends before
\texttt{optimizer.step()}.

The metric reports active allocations managed by PyTorch's CUDA allocator. It
excludes reserved-but-unused allocator blocks, the CUDA context, driver-managed
allocations, and memory allocated outside PyTorch. Optimizer state is excluded
from the measurement; Adam would otherwise add approximately two
parameter-sized buffers to every backend. All compared backends use the same
network, grid, data type, and field values.

\section{Experimental Details and Full Results}
\label{app:experiments}

\subsection{Training and Measurement Configuration}
\label{app:config}

All experiments use the following setup:
\begin{itemize}
    \item \textbf{Hardware and software:}
    one NVIDIA A100-SXM4-40GB GPU, CUDA 12.x, Python 3.10,
    PyTorch 2.9.1, and Triton 3.5.1.

    \item \textbf{Precision:}
    training uses \texttt{float32}; numerical verification uses
    \texttt{float64}.

    \item \textbf{Optimizer:}
    full-batch Adam with learning rate $10^{-3}$ and default PyTorch
    $\beta$ values.

    \item \textbf{MLP models:}
    all MLPs use Tanh activations. The per-case configurations are:

\begin{small}
\begin{tabular}{lrrrr}
\toprule
\textbf{Case} & \textbf{Input dim} & \textbf{Width} & \textbf{Hidden layers} & \textbf{Output fields} \\
\midrule
\texttt{burgers\_1d\_steady}   & 2 & 64  & 4 & 1 \\
\texttt{burgers\_1d\_unsteady} & 2 & 64  & 4 & 1 \\
\texttt{diffusion\_1d}         & 2 & 64  & 4 & 1 \\
\texttt{sod\_1d}               & 2 & 256 & 4 & 3 \\
\texttt{ldc\_2d}               & 2 & 128 & 5 & 3 \\
\texttt{tgv\_2d}               & 3 & 128 & 6 & 3 \\
\texttt{poisson\_2d}           & 2 & 128 & 5 & 1 \\
\texttt{transport\_2d}         & 3 & 128 & 5 & 1 \\
\texttt{diffusion\_2d}         & 3 & 128 & 5 & 1 \\
\texttt{heat\_2d}              & 3 & 128 & 5 & 1 \\
\texttt{wave\_2d}              & 3 & 128 & 5 & 1 \\
\texttt{allen\_cahn}           & 2 & 128 & 5 & 1 \\
\texttt{ldc\_3d}               & 3 & 128 & 5 & 4 \\
\texttt{tgv\_3d}               & 4 & 256 & 4 & 4 \\
\texttt{tgv\_3d\_smooth}       & 4 & 256 & 4 & 4 \\
\texttt{poisson\_3d}           & 3 & 128 & 5 & 1 \\
\texttt{heat\_3d}              & 4 & 128 & 5 & 1 \\
\bottomrule
\end{tabular}
\end{small}

    \item \textbf{CNN models:}
    CNN experiments use a four-layer convolutional encoder with 32 channels,
    kernel size~5, padding~2, and Tanh activations. The convolution dimension
    matches the spatial dimension of each case: 1D cases use \texttt{Conv1d},
    2D cases use \texttt{Conv2d}, and 3D cases use \texttt{Conv3d}. The input
    channel count equals the number of coordinate inputs (1--4 depending on
    dimension and time dependence); the output channel count equals the number
    of PDE fields (1--4).

    \item \textbf{\texttt{torch.compile} baseline:}
    \texttt{torch.compile(model, mode="default", fullgraph=False,
    dynamic=False)} is applied only to the neural network. The FD residual
    remains in eager PyTorch.

    \item \textbf{Warm forward--backward benchmark:}
    each timing run uses 100 warm-up iterations followed by 2,900 measured
    iterations. Five runs are performed and the median latency is reported.
    The timed region includes model forward evaluation, the PDE residual, loss,
    and backward propagation, but excludes \texttt{optimizer.step()}.

    \item \textbf{Time-to-solution benchmark:}
    T2S is recorded at the first threshold crossing, subject to a maximum budget
    of 300,000 epochs. Some runs continue after the crossing to collect additional
    convergence data; consequently, the reported total epoch count may exceed the
    epoch at which T2S is recorded. The timed region includes optimizer execution
    and one-time compilation or autotuning overhead. Triton compilation and
    autotuning take approximately 2--5\,s in the evaluated environment and are
    included in the first step; the initial \texttt{torch.compile} cost is
    treated in the same way.

    \item \textbf{Seeds:}
    \texttt{eager\_fd} and FlashPDE use five paired seeds, 42--46.
    \texttt{torch.manual\_seed},
    \texttt{torch.cuda.manual\_seed\_all}, and
    \texttt{np.random.seed} are called before model construction so paired runs
    begin from the same initialization and sampled data. The
    \texttt{vanilla} and \texttt{compile} T2S results use one seed.

    \item \textbf{Synchronization:}
    \texttt{torch.cuda.synchronize()} is called at timing boundaries so
    asynchronous GPU work is included in wall-clock measurements.
\end{itemize}

\subsection{Convergence Thresholds}
\label{app:thresholds}

Thresholds are calibrated in pilot runs and fixed before timed evaluation. The
main six cases use thresholds reached reliably by the non-vanilla backends. The
extended evaluation also includes diagnostic cases in which one or more
backends fail to reach the selected threshold.

\begin{table}[t]
\centering
\caption{PDE-loss thresholds used in the MLP time-to-solution experiments.}
\label{tab:thresholds}
\small
\begin{tabular}{lr}
\toprule
\textbf{Case} & \textbf{Threshold} \\
\midrule
\texttt{burgers\_1d\_steady}   & $10^{-5}$ \\
\texttt{burgers\_1d\_unsteady} & $10^{-3}$ \\
\texttt{diffusion\_1d}         & $2\times10^{-2}$ \\
\texttt{sod\_1d}               & $10^{-1}$ \\
\texttt{ldc\_2d}               & $8\times10^{-2}$ \\
\texttt{tgv\_2d}               & $10^{-4}$ \\
\texttt{poisson\_2d}           & $8\times10^{-2}$ \\
\texttt{transport\_2d}         & $10^{-3}$ \\
\texttt{diffusion\_2d}         & $10^{-3}$ \\
\texttt{heat\_2d}              & $10^{-3}$ \\
\texttt{wave\_2d}              & $2\times10^{-3}$ \\
\texttt{allen\_cahn}           & $10^{-4}$ \\
\texttt{ldc\_3d}               & $4.2\times10^{-1}$ \\
\texttt{tgv\_3d}               & $3.5\times10^{-3}$ \\
\texttt{tgv\_3d\_smooth}       & $4\times10^{-3}$ \\
\texttt{poisson\_3d}           & $2\times10^{-1}$ \\
\texttt{heat\_3d}              & $5\times10^{-3}$ \\
\bottomrule
\end{tabular}
\end{table}

\subsection{Main-Case Accuracy Metrics}
\label{app:error_metrics}

For cases with an exact or numerical reference field, relative error is
$\|\mathbf{q}_{\mathrm{pred}}-\mathbf{q}_{\mathrm{ref}}\|_2/
\|\mathbf{q}_{\mathrm{ref}}\|_2$. The evaluated fields and reference data
are case-specific:

\begin{table}[t]
\centering
\caption{Definitions of the reported main-case accuracy metrics.}
\label{tab:error_metrics}
\small
\begin{tabular}{lp{0.66\columnwidth}}
\toprule
\textbf{Case} & \textbf{Metric and reference} \\
\midrule
1D Burgers & Full field; numerical reference solution \\
2D Diffusion & Full scalar field; analytic solution \\
2D LDC & Centerline $u$; Ghia benchmark data \\
2D TGV & Combined full-field $u,v$; analytic solution \\
3D LDC & PDE residual-loss proxy; no reference solution \\
3D TGV & Combined $u,v,w,p$ at $t=0$; analytic initial condition \\
\bottomrule
\end{tabular}
\end{table}

The 3D LDC value is not a solution relative error, and the 3D TGV value measures
initial-condition reproduction rather than full-trajectory accuracy.

\subsection{Complete Main-Case Time-to-Solution Results}
\label{app:main_t2s}

For \texttt{eager\_fd} and FlashPDE, epochs, effective milliseconds per step, and
reported error metrics are medians over five seeds; T2S is reported as median
$\pm$ standard deviation. The effective milliseconds per step include optimizer
execution and amortized cold-start overhead. The \texttt{vanilla} and
\texttt{compile} rows use one seed.

\begin{table*}[t]
\centering
\caption{
Complete time-to-solution and convergence results for the six main cases.
Speedup is relative to the median \texttt{eager\_fd} T2S.
}
\label{tab:main_t2s_full}
\small
\setlength{\tabcolsep}{4pt}
\begin{tabular}{llrrrrr}
\toprule
\textbf{Case (threshold)}
& \textbf{Backend}
& \textbf{Epochs}
& \textbf{ms/step}
& \textbf{T2S (s)}
& \textbf{Speedup}
& \textbf{Reported error} \\
\midrule
\textbf{1D Burgers}
& \texttt{vanilla}
& 14,468 & 10.30 & 149.2 & $0.33\times$ & 0.12\% \\
($10^{-3}$)
& \texttt{eager\_fd}
& 15,189 & 3.16 & $48.5\pm31.1$ & Base & 0.51\% \\
&
\texttt{compile}
& 68,501 & 3.06 & 212.0 & $0.23\times$ & 0.09\% \\
&
\texttt{FlashPDE}
& 15,301 & 3.01 & $47.8\pm29.3$ & $1.01\times$ & 0.45\% \\
\midrule
\textbf{2D Diffusion}
& \texttt{vanilla}
& 31,583 & 31.87 & 1004.9 & $0.13\times$ & 1.90\% \\
($10^{-3}$)
& \texttt{eager\_fd}
& 25,212 & 5.29 & $133.3\pm11.6$ & Base & 2.02\% \\
&
\texttt{compile}
& 19,460 & 4.76 & 94.8 & $1.41\times$ & 1.90\% \\
&
\texttt{FlashPDE}
& 22,059 & 4.82 & $109.7\pm20.5$ & $1.22\times$ & 2.07\% \\
\midrule
\textbf{2D LDC}
& \texttt{vanilla}
& 240,117 & 17.10 & 4105.9 & $0.02\times$ & 7.13\% \\
($8\times10^{-2}$)
& \texttt{eager\_fd}
& 13,189 & 5.92 & $76.9\pm34.2$ & Base & 4.93\% \\
&
\texttt{compile}
& 99,568 & 6.04 & 591.0 & $0.13\times$ & 1.77\% \\
&
\texttt{FlashPDE}
& 13,784 & 3.35 & $43.9\pm10.6$ & $1.75\times$ & 4.66\% \\
\midrule
\textbf{2D TGV}
& \texttt{vanilla}
& 180,659 & 78.78 & 14233.8 & $0.07\times$ & 1.60\% \\
($10^{-4}$)
& \texttt{eager\_fd}
& 106,244 & 9.68 & $1026.0\pm144.2$ & Base & 3.24\% \\
&
\texttt{compile}
& 129,304 & 8.75 & 1136.2 & $0.90\times$ & 3.31\% \\
&
\texttt{FlashPDE}
& 107,259 & 6.49 & $696.1\pm25.6$ & $1.47\times$ & 3.27\% \\
\midrule
\textbf{3D LDC}
& \texttt{vanilla}
& --- & OOM & --- & --- & --- \\
($4.2\times10^{-1}$)
& \texttt{eager\_fd}
& 18,613 & 14.36 & $268.4\pm51.9$ & Base & 0.26\% \\
&
\texttt{compile}
& 13,873 & 12.91 & 186.5 & $1.44\times$ & 0.017\% \\
&
\texttt{FlashPDE}
& 15,174 & 7.66 & $116.6\pm29.6$ & $2.30\times$ & 0.32\% \\
\midrule
\textbf{3D TGV}
& \texttt{vanilla}
& --- & OOM & --- & --- & --- \\
($3.5\times10^{-3}$)
& \texttt{eager\_fd}
& 20,785 & 41.64 & $866.7\pm68.5$ & Base & 3.03\% \\
&
\texttt{compile}
& 18,775 & 24.20 & 454.0 & $1.91\times$ & 2.93\% \\
&
\texttt{FlashPDE}
& 20,284 & 39.37 & $805.0\pm68.8$ & $1.08\times$ & 3.08\% \\
\bottomrule
\end{tabular}
\end{table*}

\subsection{Extended MLP Results}
\label{app:track_results}

\subsubsection{Extended Forward--Backward Throughput}

\begin{table*}[t]
\centering
\caption{
Extended forward--backward latency in milliseconds. Speedup compares FlashPDE
with eager PyTorch FD.
}
\label{tab:app_microbenchmark}
\small
\begin{tabular}{lrrrrr}
\toprule
\textbf{Case}
& \textbf{vanilla}
& \textbf{eager FD}
& \textbf{compile}
& \textbf{FlashPDE}
& \textbf{Speedup} \\
\midrule
1D Steady Burgers & 4.35 & 1.98 & 2.35 & 1.99 & $0.99\times$ \\
Allen--Cahn        & 6.01 & 1.48 & ---  & 1.20 & $1.23\times$ \\
1D Sod Shock Tube & 104.28 & 11.82 & 9.96 & 11.57 & $1.02\times$ \\
1D Heat Equation  & 5.10 & 2.43 & 2.64 & 2.33 & $1.04\times$ \\
2D Scalar Transport & 31.66 & 3.62 & 3.81 & 3.05 & $1.19\times$ \\
2D Poisson        & 8.49 & 2.89 & 3.06 & 2.41 & $1.20\times$ \\
2D Heat Equation  & 31.68 & 3.64 & 3.98 & 3.08 & $1.18\times$ \\
2D Wave Equation  & 43.36 & 4.00 & 4.33 & 3.32 & $1.20\times$ \\
3D Poisson        & 414.40 & 28.08 & 22.36 & 27.67 & $1.01\times$ \\
3D TGV Smooth     & --- & 24.16 & 20.68 & 21.95 & $1.10\times$ \\
3D Heat Equation  & 194.55 & 15.75 & 16.86 & 17.24 & $0.91\times$ \\
\bottomrule
\end{tabular}
\end{table*}

The extended cases show that FlashPDE's full-step benefit is
workload-dependent. It improves eager FD latency on eight cases, is approximately
neutral on 1D steady Burgers, 1D Sod, and 3D Poisson, and is slower on 3D heat.

\subsubsection{Extended Time-to-Solution Results}

In Table~\ref{tab:app_end2end}, \emph{Epochs} denotes the total number of
executed epochs, whereas T2S is the time of the first threshold crossing and may
therefore occur before the final reported epoch. A speedup relative to
\texttt{eager\_fd} is shown only when that backend reaches the threshold.

\begin{table*}[t]
\centering
\caption{
Extended MLP time-to-solution results. ``Div.'' denotes numerical divergence;
``---'' denotes an unavailable or undefined metric.
}
\label{tab:app_end2end}
\small
\setlength{\tabcolsep}{4pt}
\begin{tabular}{llrrrr}
\toprule
\textbf{Case (threshold)}
& \textbf{Backend}
& \textbf{Epochs}
& \textbf{T2S (s)}
& \textbf{Speedup}
& \textbf{Error} \\
\midrule
\textbf{1D Steady Burgers}
& \texttt{vanilla}
& 2,674 & 12.2 & $0.47\times$ & 31.9\% \\
($10^{-5}$)
& \texttt{eager\_fd}
& 2,601 & 5.7 & Base & 50.9\% \\
&
\texttt{compile}
& 2,620 & 7.2 & $0.79\times$ & 44.1\% \\
&
\texttt{FlashPDE}
& 2,574 & 6.3 & $0.90\times$ & 39.1\% \\
\midrule
\textbf{Allen--Cahn}
& \texttt{vanilla}
& 13,671 & 212.0 & $0.92\times$ & --- \\
($10^{-4}$)
& \texttt{eager\_fd}
& 69,454 & 194.7 & Base & $6.56\times10^{-5}$ \\
&
\texttt{compile}
& 67,172 & 521.0 & $0.37\times$ & --- \\
&
\texttt{FlashPDE}
& 62,626 & 176.6 & $1.10\times$ & $7.60\times10^{-5}$ \\
\midrule
\textbf{2D Scalar Transport}
& \texttt{vanilla}
& 37,181 & 1184.2 & $0.13\times$ & 4.90\% \\
($10^{-3}$)
& \texttt{eager\_fd}
& 32,037 & 154.2 & Base & 4.90\% \\
&
\texttt{compile}
& 33,750 & 153.1 & $1.01\times$ & 4.95\% \\
&
\texttt{FlashPDE}
& 29,195 & 106.6 & $1.45\times$ & 4.93\% \\
\midrule
\textbf{1D Sod Shock}
& \texttt{vanilla}
& --- & --- & --- & Div. \\
($10^{-1}$)
& \texttt{eager\_fd}
& --- & --- & --- & Div. \\
&
\texttt{compile}
& --- & --- & --- & Div. \\
&
\texttt{FlashPDE}
& 141,990 & 1608.7 & --- & 49.9\% \\
\midrule
\textbf{2D Poisson}
& \texttt{vanilla}
& 1,531 & 13.1 & $43.9\times$ & 0.17\% \\
($8\times10^{-2}$)
& \texttt{eager\_fd}
& 197,646 & 575.7 & Base & 2.25\% \\
&
\texttt{compile}
& 64,012 & 193.3 & $2.98\times$ & 0.24\% \\
&
\texttt{FlashPDE}
& 190,459 & 495.8 & $1.16\times$ & 0.17\% \\
\midrule
\textbf{3D Poisson}
& \texttt{vanilla}
& 126,667 & 233.0 & $1.50\times$ & 0.02\% \\
($2\times10^{-1}$)
& \texttt{eager\_fd}
& 300,000 & 350.0 & Base & 0.14\% \\
&
\texttt{compile}
& 300,000 & 294.0 & $1.19\times$ & 0.14\% \\
&
\texttt{FlashPDE}
& 300,000 & 345.0 & $1.01\times$ & 0.14\% \\
\midrule
\textbf{1D Heat Equation}
& \texttt{vanilla}
& 647 & 3.4 & $32.1\times$ & 0.04\% \\
($2\times10^{-2}$)
& \texttt{eager\_fd}
& 40,941 & 109.2 & Base & 0.76\% \\
&
\texttt{compile}
& 1,090 & 3.0 & $36.4\times$ & 0.08\% \\
&
\texttt{FlashPDE}
& 53,065 & 130.5 & $0.84\times$ & 0.78\% \\
\midrule
\textbf{3D TGV Smooth}
& \texttt{vanilla}
& --- & --- & --- & --- \\
($4\times10^{-3}$)
& \texttt{eager\_fd}
& 19,731 & 574.3 & Base & 2.93\% \\
&
\texttt{compile}
& 21,204 & 540.1 & $1.06\times$ & 2.93\% \\
&
\texttt{FlashPDE}
& 13,773 & 314.0 & $1.83\times$ & 2.93\% \\
\midrule
\textbf{2D Heat Equation}
& \texttt{vanilla}
& 37,354 & 1189 & $0.17\times$ & 7.78\% \\
($10^{-3}$)
& \texttt{eager\_fd}
& 47,128 & 198 & Base & 11.51\% \\
&
\texttt{compile}
& 47,631 & 191 & $1.04\times$ & 11.88\% \\
&
\texttt{FlashPDE}
& 49,035 & 177 & $1.12\times$ & 11.88\% \\
\midrule
\textbf{2D Wave Equation}
& \texttt{vanilla}
& 21,166 & 922 & $0.29\times$ & 19.26\% \\
($2\times10^{-3}$)
& \texttt{eager\_fd}
& 59,872 & 268 & Base & 21.96\% \\
&
\texttt{compile}
& 61,129 & 261 & $1.03\times$ & 21.83\% \\
&
\texttt{FlashPDE}
& 61,862 & 243 & $1.10\times$ & 21.79\% \\
\midrule
\textbf{3D Heat Equation}
& \texttt{vanilla}
& 27,065 & 4411 & $0.19\times$ & 8.87\% \\
($5\times10^{-3}$)
& \texttt{eager\_fd}
& 33,195 & 840 & Base & 7.89\% \\
&
\texttt{compile}
& 31,309 & 767 & $1.10\times$ & 8.91\% \\
&
\texttt{FlashPDE}
& 33,736 & 848 & $0.99\times$ & 9.11\% \\
\bottomrule
\end{tabular}
\end{table*}

\subsection{Additional Kernel-Scaling Results}
\label{app:scaling}

\begin{figure*}[t]
\centering
\includegraphics[width=\textwidth]{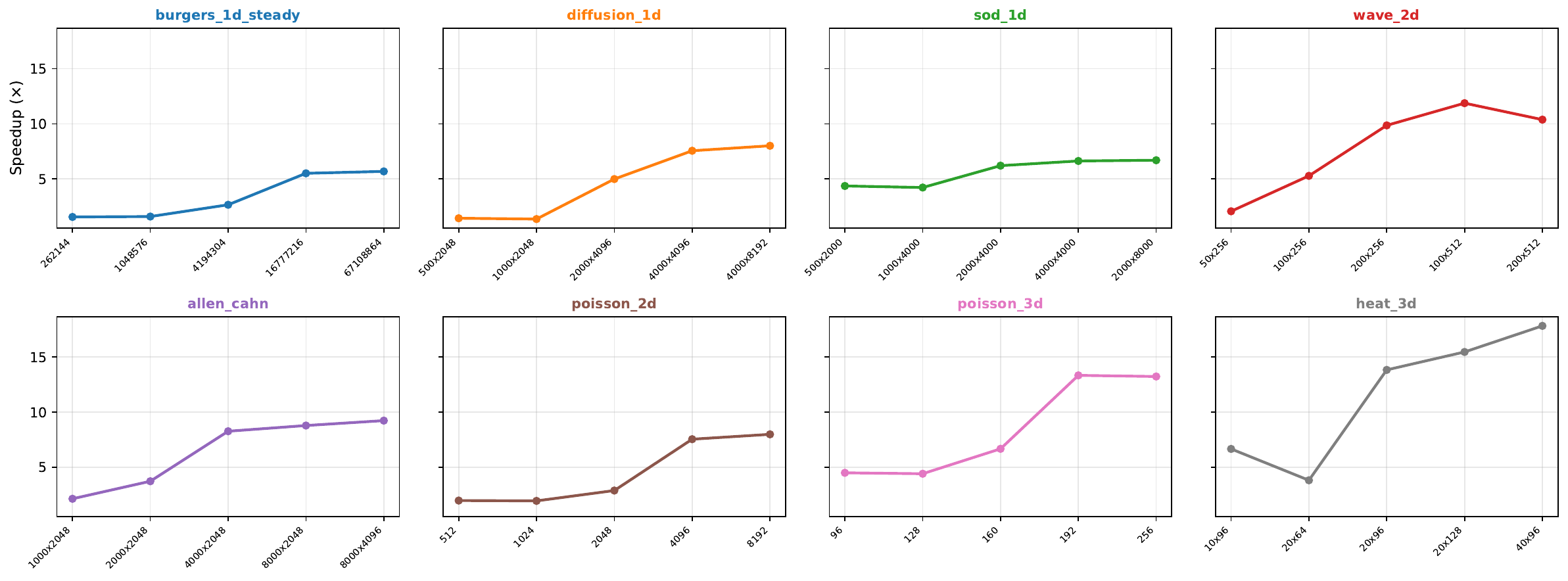}
\caption{
Kernel-only FlashPDE speedup over eager PyTorch FD for the eight additional
operators. Most cases show an overall increase with grid size, although several
trajectories are non-monotonic at intermediate or maximum evaluated sizes.
}
\label{fig:scaling_appendix}
\end{figure*}

Figure~\ref{fig:scaling_appendix} reports kernel-only speedups for the eight
operators not displayed in the main scaling figure.

\subsection{CNN Results}
\label{app:cnn_track2}

The same FlashPDE operators are used with CNN field generators without changing
the operator implementations. These experiments do not propose a new CNN
architecture; instead, they test whether FlashPDE can be reused when the field
generator is changed from an MLP to a convolutional model.

\subsubsection{CNN Forward--Backward Throughput}

\begin{table*}[t]
\centering
\caption{
CNN forward--backward latency and peak allocated memory. Each backend entry is
milliseconds per step / GB.
}
\label{tab:cnn_microbenchmark}
\small
\begin{tabular}{lrrrr}
\toprule
\textbf{Case}
& \textbf{CNN eager FD}
& \textbf{cnn\_compile}
& \textbf{cnn\_FlashPDE}
& \textbf{Speedup} \\
\midrule
1D Steady Burgers & 2.11 / 0.018 & 2.26 / 0.018 & 1.85 / 0.018 & $1.14\times$ \\
1D Heat Equation & 2.55 / 0.024 & 2.79 / 0.024 & 2.38 / 0.024 & $1.07\times$ \\
1D Sod Shock Tube & 5.63 / 0.178 & 5.86 / 0.180 & 3.36 / 0.180 & $1.68\times$ \\
2D Scalar Transport & 4.22 / 0.244 & 4.30 / 0.244 & 3.79 / 0.244 & $1.11\times$ \\
2D Poisson & 2.67 / 0.066 & 2.85 / 0.066 & 2.48 / 0.066 & $1.08\times$ \\
2D Heat Equation & 4.12 / 0.244 & 3.98 / 0.244 & 3.84 / 0.244 & $1.07\times$ \\
2D Wave Equation & 4.15 / 0.244 & 4.51 / 0.244 & 3.91 / 0.244 & $1.07\times$ \\
3D Poisson & 75.36 / 0.251 & 74.21 / 0.252 & 74.44 / 0.252 & $1.01\times$ \\
3D TGV Smooth & 31.83 / 0.307 & 31.55 / 0.313 & 29.65 / 0.311 & $1.07\times$ \\
3D Heat Equation & 44.15 / 0.290 & 43.52 / 0.292 & 43.48 / 0.291 & $1.02\times$ \\
\bottomrule
\end{tabular}
\end{table*}

\subsubsection{CNN Time-to-Solution}

Some CNN runs are terminated after the loss plateaus. A dash in the T2S column
indicates that the selected threshold was not reached.

\begin{table*}[t]
\centering
\caption{
CNN convergence results for eight additional configurations.
}
\label{tab:cnn_end2end}
\small
\setlength{\tabcolsep}{4pt}
\begin{tabular}{llrrr}
\toprule
\textbf{Case (threshold)}
& \textbf{Backend}
& \textbf{Epochs}
& \textbf{T2S (s)}
& \textbf{Error} \\
\midrule
\textbf{1D Steady Burgers}
& CNN eager FD
& 218,510 & 460.6 & 0.243 \\
($10^{-5}$)
& \texttt{cnn\_compile}
& 212,581 & 502.4 & 0.425 \\
&
\texttt{cnn\_FlashPDE}
& 201,469 & 381.4 & 0.373 \\
\midrule
\textbf{1D Heat Equation}
& CNN eager FD
& 171,363 & 444.6 & 0.17\% \\
($2\times10^{-2}$)
& \texttt{cnn\_compile}
& 168,873 & 481.9 & 0.18\% \\
&
\texttt{cnn\_FlashPDE}
& 178,831 & 428.8 & 0.19\% \\
\midrule
\textbf{1D Sod Shock}
& CNN eager FD
& 36,221 & 225.6 & 40.3\% \\
($5\times10^{-2}$)
& \texttt{cnn\_compile}
& 52,434 & 326.4 & 41.0\% \\
&
\texttt{cnn\_FlashPDE}
& 41,071 & 192.0 & --- \\
\midrule
\textbf{2D Scalar Transport}
& CNN eager FD
& 75,179 & 445.3 & --- \\
($10^{-3}$)
& \texttt{cnn\_compile}
& 101,607 & 604.8 & --- \\
&
\texttt{cnn\_FlashPDE}
& 76,693 & 355.0 & --- \\
\midrule
\textbf{2D Poisson}
& CNN eager FD
& 300,000 & --- & 0.07\% \\
($8\times10^{-2}$)
& \texttt{cnn\_compile}
& 300,000 & --- & 0.21\% \\
&
\texttt{cnn\_FlashPDE}
& 300,000 & --- & 0.18\% \\
\midrule
\textbf{2D Heat Equation}
& CNN eager FD
& 129,169 & --- & 2.79\% \\
($10^{-3}$)
& \texttt{cnn\_compile}
& 79,018 & --- & 2.86\% \\
&
\texttt{cnn\_FlashPDE}
& 31,786 & --- & 2.92\% \\
\midrule
\textbf{2D Wave Equation}
& CNN eager FD
& 300,000 & --- & 7.30\% \\
($2\times10^{-3}$)
& \texttt{cnn\_compile}
& 300,000 & --- & 6.01\% \\
&
\texttt{cnn\_FlashPDE}
& 300,000 & --- & 7.72\% \\
\midrule
\textbf{Allen--Cahn}
& CNN eager FD
& 24,033 & 71.1 & $1.92\times10^{-4}$ \\
($10^{-4}$)
& \texttt{cnn\_compile}
& --- & --- & --- \\
&
\texttt{cnn\_FlashPDE}
& 25,042 & 63.3 & $2.84\times10^{-4}$ \\
\bottomrule
\end{tabular}
\end{table*}

\newpage

\end{document}